\documentclass[lettersize,journal]{IEEEtran}
\usepackage{amsmath,amsfonts}
\usepackage{algorithmic}
\usepackage{algorithm}
\usepackage{array}
\usepackage[caption=false,font=normalsize,labelfont=sf,textfont=sf]{subfig}
\usepackage{textcomp}
\usepackage{stfloats}
\usepackage{url}
\usepackage{verbatim}
\usepackage{graphicx}
\usepackage{cite}

\usepackage{booktabs}
\usepackage{multirow}
\usepackage{caption}
\usepackage[table]{xcolor}
\definecolor{revisionred}{RGB}{180,30,30}
\usepackage{pifont}
\usepackage{amssymb} 

\begin{document}

\title{Partition, Refine, and Detect: Preserving Novel Category Capacity in Open-Vocabulary Object Detection}

\author{Wei Zhang, Zhu Teng, Zhixiang Chi, Junnan Zou, Yi Liu, Jianping Fan
        % <-this % stops a space
\thanks{Wei Zhang, Zhu Teng and Junnan Zou are with the Beijing Jiaotong University, Beijing 100044, China (E-mail: {wzhang0926, zteng, zoujunnan}@bjtu.edu.cn). Zhixiang Chi is with University of Toronto, Toronto, ON M5S 2E8, Canada (E-mail: zhxchi@ece.utoronto.ca). Yi Liu is with Baidu Inc., Shenzhen, Guangdong 518000, China (E-mail: liuyi22@baidu.com). Jianping Fan is with institute of Intelligent Robotics and Advanced Manufacturing, Fudan University, Shanghai, China (E-mail: jianpingfan@fudan.edu.cn). \textit{(Corresponding author: Zhu Teng)}}

\thanks{This work was supported by the Fundamental Research Funds for the Central Universities (2026JBZD027) and the Natural Science Foundation of China (Grant NO. 62472025).}
%\thanks{\textsuperscript{*}Corresponding author.}
}
\markboth{Journal of \LaTeX\ Class Files,~Vol.~14, No.~8, August~2021}%
{Shell \MakeLowercase{\textit{et al.}}: A Sample Article Using IEEEtran.cls for IEEE Journals}



\maketitle

\begin{abstract}
Open-vocabulary object detection (OVD) expects the model trained on base classes to recognize both base and novel classes at inference. Typical methods rely on CLIP-style vision–language embeddings, yet the learned space is often biased toward base categories, with limited effective capacity preserved for novel concepts. Such a base-dominated representation can compress novel semantics, leading to semantic collapse and mis-calibrated confidence at inference.
To address this issue, we present PR-OVD, a partition-and-refine framework that explicitly reserves capacity for novel categories and strengthens category separation, while being seamlessly integrated into different detector architectures. Specifically, PR-OVD first performs coarse partitioning to build global divisions for base, novel, and an explicit background partition in both visual and textual modalities, preventing object-irrelevant embeddings from leaking into category-specific representations. It then applies category-level refinement within each partition to sharpen decision boundaries and alleviate the lack of fine-grained region-level pretraining in CLIP.
To further provide reliable visual anchors for novel categories, we design a cognitively inspired Structure Prompt (SP) that encodes both egocentric attributes and allocentric context, facilitating stable partition-and-refine learning. Extensive experiments on two OVD benchmarks show that PR-OVD delivers an average 6.5\% proportional improvement on novel categories over the state-of-the-art methods. Notably, PR-OVD also benefits other open-vocabulary tasks, including segmentation and classification, further confirming its effectiveness and generality.

\end{abstract}

\begin{IEEEkeywords}
Open Vocabulary Object Detection, Representation Learning, Prompt Learning.
\end{IEEEkeywords}
\section{Introduction}
\label{sec:intro}

Traditional vision methods \cite{6_mrcnn,8_yolo} rely on a fixed set of predefined categories, limiting their applicability in real-world scenarios where new concepts frequently appear. Open-vocabulary understanding addresses this limitation by leveraging language information, such as category names or image captions. This paradigm has been applied to various tasks, including segmentation \cite{82_tmmseg, chi1}, object detection \cite{33_clipself}, video understanding \cite{83_tmmvideo,84_tmmvideo2}, and autonomous driving \cite{85_tmmautodive}. In this work, we focus on Open-Vocabulary Object Detection (OVD) \cite{10_ovrcnn}, which extends conventional paradigm to recognize objects whose categories are unseen during training. Unlike closed-vocabulary detection, where all categories are learned with instance-level annotations, OVD aims to recognize novel categories without corresponding visual supervision by leveraging semantic knowledge from language.

\begin{figure}[t]
\centering
\includegraphics[width=\columnwidth]{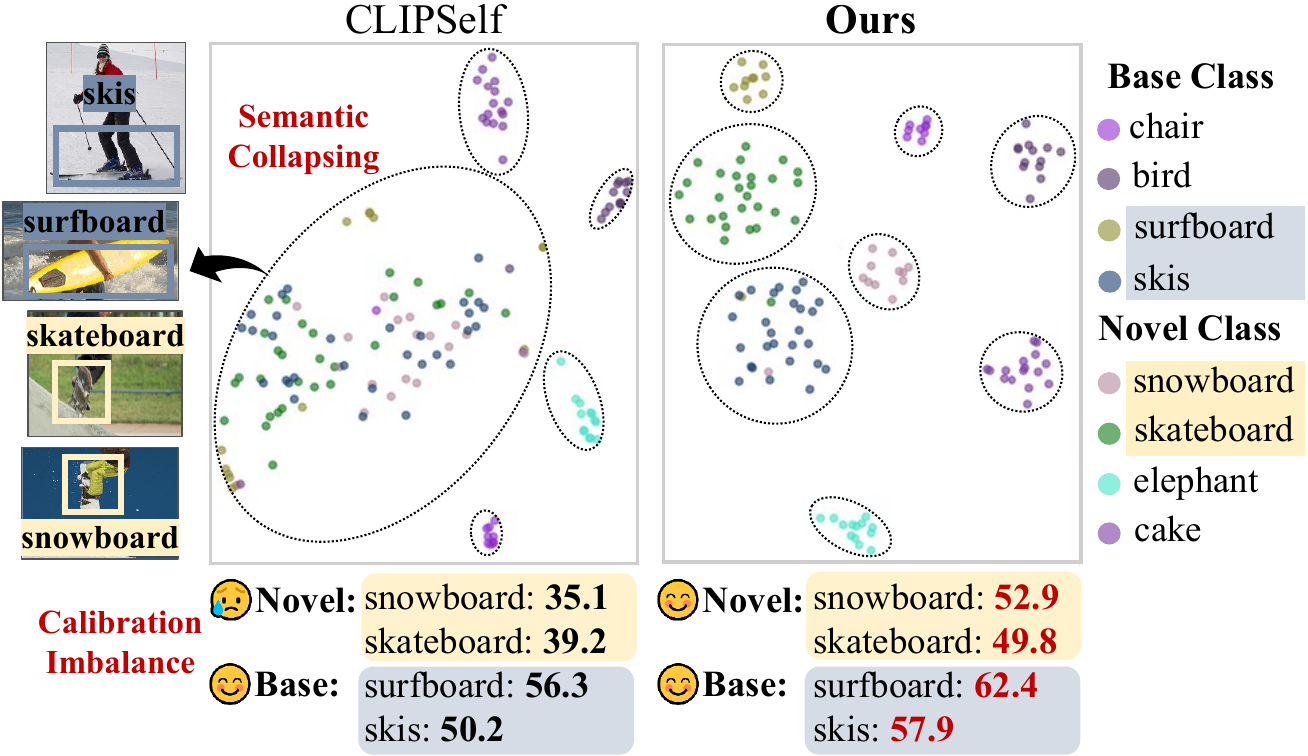}
\caption{\textbf{Semantic collapsing and calibration imbalance in OVD.} 
\textbf{\textit{Left:}} t-SNE visualization and per-class performance on OV-COCO of a SoTA OVD method (CLIPSelf \cite{33_clipself}). Semantically similar base and novel categories are entangled, resulting in collapsed decision boundaries and calibration imbalance. 
\textbf{\textit{Right:}} Our PR-OVD explicitly reserves representational space for novel categories, forming clear base–novel decision boundaries and substantially improving novel performance.}
\label{fig:intro}
\vspace{-11pt}
\end{figure}

% \begin{wrapfigure}{r}{0.6\textwidth}
% \centering
% \includegraphics[width=0.6\textwidth]{figs/intro.pdf}
% \caption{\textbf{Semantic collapsing and calibration imbalance in OVD.} 
% \textbf{\textit{Left:}} t-SNE visualization and per-class performance on OV-COCO of a SoTA OVD method (CLIPSelf \cite{33_clipself}). Semantically similar base and novel categories are entangled, resulting in collapsed decision boundaries and calibration imbalance. 
% \textbf{\textit{Right:}} Our PR-OVD explicitly reserves representational space for novel categories, forming clear base–novel decision boundaries and substantially improving novel performance.}
% \label{fig:intro}
% \vspace{-20pt}
% \end{wrapfigure}

Most existing OVD methods optimize detection loss using only base categories during training, while at inference time the model is required to recognize both base and novel categories \cite{51_cake, 33_clipself}. \textbf{\textit{This misalignment between the training objective and the testing protocol introduces a fundamental optimization inconsistency.}} As training progresses, the embedding space is progressively reshaped to maximize separability among base classes only. Without inter-class constraints, nearby base categories tend to spread apart to occupy larger regions of the embedding space, implicitly compressing the representational capacity available to novel categories.
At inference time, this imbalance manifests as a clear semantic collapsing phenomenon as shown in the left of Fig.\ref{fig:intro}: novel category clusters are absorbed into semantically similar base clusters, lacking sufficient geometric margin. As a result, base predictions become over-confident, while novel categories consistently receive lower confidence scores, leading to severe calibration imbalance.

To address this, we propose the Partition-and-Refine OVD (PR-OVD). Its core insight is to explicitly partition the embedding space to reserve dedicated regions for categories unseen in training but required at test time, preventing semantic absorption by dominant base classes. Within each partition, category-level refinement is then applied to further reinforce inter-category constraints. PR-OVD establishes clear decision boundaries between base and novel categories through an architecture-agnostic framework (as shown in Fig. \ref{fig:intro}, right), effectively mitigating semantic collapse and calibration imbalance. This design is inspired by human visual cognition \cite{72_human1,73_human2}. Humans perceive objects hierarchically: they first partition them into \textit{known} and \textit{unknown}, then perform fine-grained recognition within each group \cite{72_human1}. Importantly, unknown objects are maintained as structured representations rather than collapsed into known categories.

Specifically, PR-OVD consists of two components:
% \textbf{\textit{I. Coarse partition}}, establishing global spatial division in both visual and textual sides. In the textual modality, we model base and novel prototypes, and additionally introduce the background prototype to reflect the prevalence of background proposals in object detection and to prevent background semantics from interfering with object-related representations. Prototype-level partition learning enforcing global partitions. 

\noindent\textbf{\textit{I. Coarse partition}} establishes a global spatial division in both visual and textual spaces by modeling base and novel prototypes, and additionally introduce a background prototype to account for the large number of background proposals inherent in object detection. Prototype-level contrastive learning is then applied to enforce clear global partitions.
In the visual modality, two challenges emerge: (1) Prototype construction: novel classes lack real visual samples. Traditional pseudo-labeling methods leverage external captions \cite{74_cc3m,75_caption} or MLLMs to mine potential novel regions, yet such supervision is inherently noisy, yielding inaccurate region-word alignment \cite{52_cckt}. We propose Structure Prompt (SP), a category-wise template that enables the LLM to generate multi-dimensional enhanced descriptions by combining intrinsic attributes with contextual cues. The descriptions then guide a generative model to synthesize a few representative novel images for visual prototype construction and subsequent category-level refinement. SP shifts from mining to generation, unlocking a controllable pathway for constructing reliable visual prototypes of novel categories. (2) Embedding learning: novel-background semantic overlap risks boundary confusion. We adopt asymmetric prototype learning in visual contrastive learning, where reliable relationships between base–novel and base–background partitions are established, while the novel–background relationship is not explicitly constrained due to unreliable supervision.
\textbf{\textit{II. Category-level Refinement}}, which performs bidirectional, region-level alignment within each \textit{partition}, reinforcing inter-class constraints and compensating for the absence of region-level pretraining in CLIP \cite{27_clip, chi2}, which is crucial for detection tasks.

Our main contributions are as follows:
\begin{itemize}
\item We identify and analyze a structural optimization bias in OVD, showing that the training–testing misalignment leads to semantic collapsing and calibration imbalance in embedding geometry.

\item We propose PR-OVD, a partition-and-refine framework inspired by human visual cognition that simultaneously reserves dedicated regions for novel categories and performs fine-grained category alignment, establishing clear base–novel decision boundaries in an architecture-agnostic manner.

\item We design a generative pseudo-labeling strategy, structure prompt (SP), to generate proxy visual features of novel categories. SP incorporates both egocentric (attributes) and allocentric (environment relations) perspectives, serving as visual prototype of novel category.

\item Beyond advancing open-vocabulary detection, our work serves as a generalizable paradigm that transfers seamlessly to other ov-tasks such as segmentation and multi-label classification, consistently delivering significant gains. Extensive experiments validate the superiority and broad applicability of our framework, offering a fresh perspective for open vocabulary understanding.

\end{itemize}

\section{Related Work}
\label{sec:rl}

\noindent\textbf{Open-Vocabulary Object Detection.}
% \textbf{Open-Vocabulary Object Detection.}
Existing OVD methods can be broadly categorized into CLIP-based and pseudo-label-based approaches.
CLIP-based methods transfer CLIP’s open-vocabulary capabilities directly into detection frameworks \cite{51_cake,14_sasdet}, or enhance CLIP with region-level perception through retraining \cite{43_rovit,44_cfmvit}.
In contrast, pseudo-label-based methods aim to mine region-word pairs for novel categories, providing supervision signals for training. For example, 
% PB-OVD \cite{12_pbovd} combines VLMs with Grad-CAM \cite{24_gradcam} to generate activation maps that highlight potential novel regions. 
% ProxyDet \cite{16_proxydet} observes that many novel categories reside within the convex hull of base classes in the CLIP embedding space \cite{27_clip}, and generates proxy features via linear interpolation \cite{26_mixup}. 
RALF \cite{15_ralf} assigns pseudo-labels by computing region-word similarity, further refining them through retrieval-based hard negative mining \cite{25_retrieval}. SAS-Det \cite{14_sasdet} decouples pseudo-label prediction for novel classes from base-class classification via a teacher-student framework.
% However, existing OVD methods suffer from semantic collapsing and calibration imbalance due to the misalignment between training and testing objectives, while pseudo-label strategies introduce inherently noisy supervision. In contrast, we propose a new OVD framework that explicitly partitions the embedding space to reserve dedicated regions for novel classes.
Moreover, ProxyDet \cite{16_proxydet} observes that novel categories often lie near the convex hull formed by base categories in the CLIP embedding space and synthesizes proxy novel classes through classwise mixup.
However, pseudo-label-based strategies introduce noisy supervision by directly optimizing detectors with generated novel samples. Different from these approaches, PR-OVD does not recover missing supervision through pseudo labels. Instead, we propose the structure prompt which serves as a structured visual-anchor construction mechanism that generates category-relevant multi-scale visual anchors for novel categories, and constructs novel-category visual anchors to explicitly reserve representation space and performs partition-and-refine learning to improve base–novel separability.

\noindent\textbf{Vision-Language Alignment in Images and Regions.}
Vision-language pre-training has led to the development of models that learn well-aligned representations between images and text, especially under unsupervised or weakly supervised settings. Early methods such as MoCo \cite{28_moco} and SimCLR \cite{29_simclr} perform instance-level discrimination on entire images, while CLIP extends it to cross-modal alignment using global image–text pairs. However, , such image-level paradigms lack the fine-grained, region-level semantics required for dense prediction, motivating recent advances in region-aware alignment. For example, DetCo \cite{30_detco} introduces self-supervised region-level objectives to capture structural relationships among image patches. DetCLIP \cite{18_detclip} constructs proposal-text contrastive pairs atop CLIP to enhance region-level cross-modal alignment. 
% However, existing methods typically rely on large-scale region-text pre-training, resulting in significant computational overhead. 
Our PR-OVD implicitly learns region-word alignment through category-level refinement. By leveraging structure-prompt-guided novel features, it achieves fine-grained engagement with minimal additional cost.

\section{Method}
\label{sec:method}

\subsection{Preliminary}

% Open-vocabulary object detection (OVD) addresses the limitations of conventional object detectors that are restricted to a closed set of predefined categories. 
Open-vocabulary object detection generalizes to novel classes by leveraging an open concept space and a class-agnostic region proposal network, reformulating the final classification stage as a region-to-text similarity task via CLIP’s cross-modal alignment.
In general, base classes $C^{\mathcal{B}}$ refer to the categories with annotations available during training, while novel classes $C^{\mathcal{N}}$ that lack such supervision during training but required to be recognized at test time ($C^{\mathcal{B}} \cap C^{\mathcal{N}}=\varnothing$). Furthermore, we follow protocols of recent OVD methods~\cite{17_survey,14_sasdet,15_ralf,52_cckt}, the category names of novel classes $C^{\mathcal{N}}$ are available as semantic information, while their instance-level annotations are removed during training. At inference time, the model needs to detect objects from $C^{\mathcal{B}} \cup C^{\mathcal{N}}$.

% Furthermore, to better reflect real-world conditions, an open vocabulary set $C^{\mathcal{O}}$ in the textual domain is also introduced. Many OVD methods also relax the constraint of requiring access to novel class vocabulary during training \cite{17_survey,14_sasdet,15_ralf,52_cckt}. At inference time, the model needs to detect objects from $C^{\mathcal{B}} \cup C^{\mathcal{N}}$.

\begin{figure*}[t]
\centering
\includegraphics[width=0.85\textwidth]{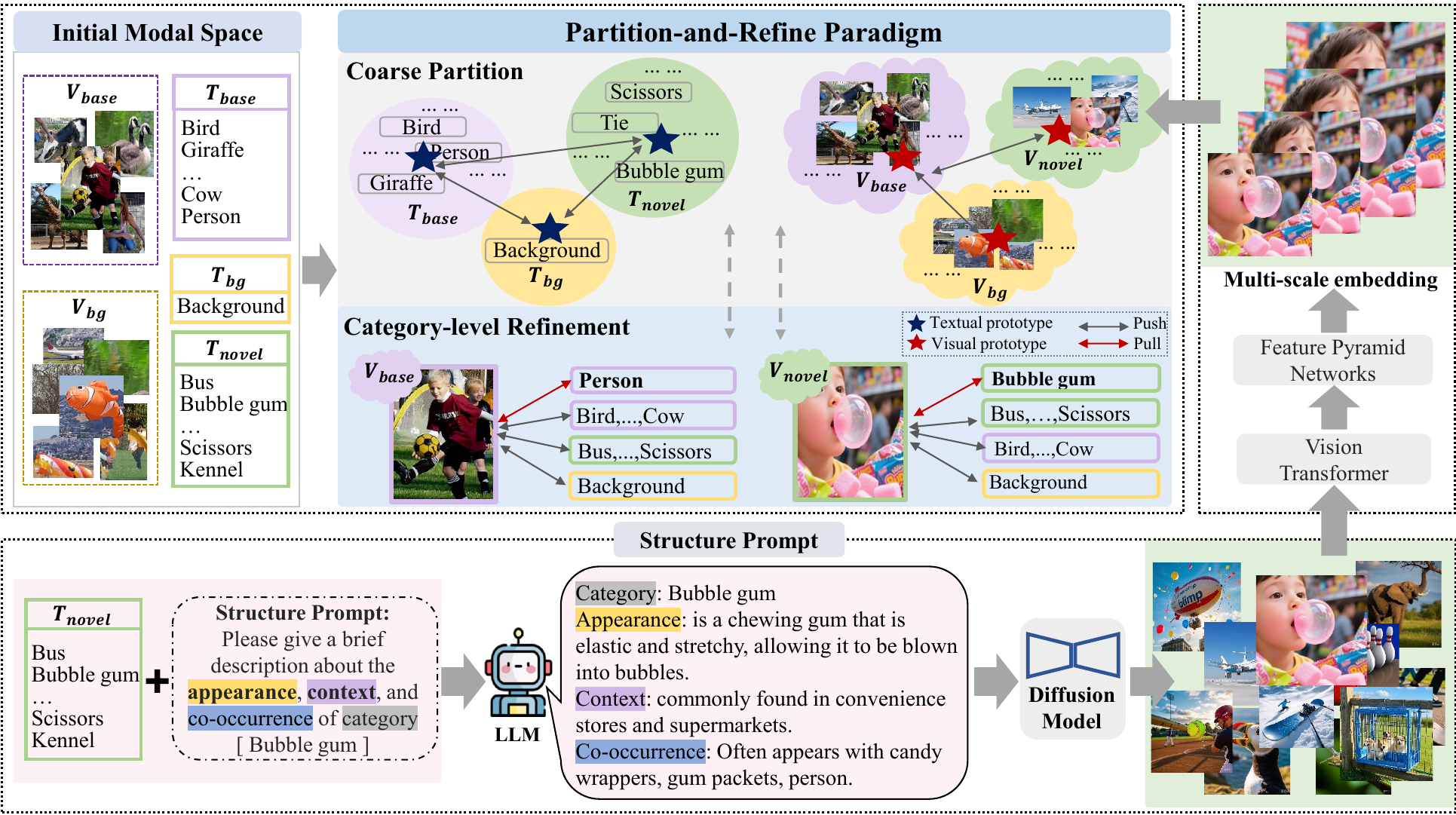}
\caption{\textbf{The architecture of our PR-OVD.}
Our partition-and-refine paradigm first establishes global partitions in visual and textual embeddings to reserve dedicated space for novel categories, then applies category-level refinement within each partition to sharpen inter-class boundaries. Among them, structure prompt (SP) encodes multi-dimensional category descriptions and guides a generative model to synthesize representative novel images, providing enriched visual anchors for both prototype construction and category-level refinement.
} 
\label{fig:method}
\vspace{-10pt}
\end{figure*}

\subsection{Overview of PR-OVD}
To mitigate the optimization misalignment between training and testing, the training process should reserve independent representational space for novel categories \cite{chi3}. Therefore, our PR-OVD adopt a partition-and-refine framework, as illustrated in Fig.\ref{fig:method}. By enforcing structural separation among base, novel, and background partitions, the model prevents novel representations from being absorbed by dominant base clusters. It also refines category alignment within these constraints to sharpen inter-category boundaries. 
To build novel-class prototype, we further design a structure prompt to support representation learning for novel categories within PR-OVD. 
% PR-OVD establishes clear base–novel separability while remaining architecture-agnostic and compatible with diverse detector architectures.

\subsection{Coarse Partition Module}
\label{sec:3.3}
This module establishes a global spatial division in both language and visual embeddings, inspired by human cognition, which first distinguishes known from unknown \cite{72_human1} before fine-grained recognition. It explicitly carves out dedicated regions for \textit{base}, \textit{novel}, and \textit{background} categories, preventing the embedding space from being monopolized by base classes and promoting separated high-level structures.

\noindent\textbf{Language space partition.}
We separate the three coarse partitions of base, novel, and background in textual space, which is inherently structured and discriminative. They provide reliable global anchors that help mitigate semantic aliasing and noise in the region-level visual space caused by occlusion, pose variation, and clutter. 
Specifically, we compute coarse-grained prototypes for each partition in the textual embedding space as follows:
$\mathbf{p}_{tp}=\frac{1}{\left|C_{tp}\right|} \sum_{c \in C_{tp}} f_{\operatorname{text}}(c)$. Here, $f_{\operatorname{text}}(\cdot)$ represents the embedding feature output by a text encoder (\textit{e.g.}, CLIP-Text), and $C_{tp}$ is the set of categories in partition $tp$ (i.e., \textit{base}, \textit{novel}, and \textit{background}). To promote inter-partition disentanglement, we introduce a partition-level textual prototype alignment objective \cite{35_procontrast} as Equ.(\ref{equ:t2t}).
{
\small
\begin{equation}\label{equ:t2t}
\mathcal{L}_{\operatorname{ct}}\left(T,\!T^{+},\!T^{-}\right)\!=\!-\!\sum_{tp}\!\log\!\frac{\exp\left(\operatorname{sim}\left(\mathbf{p}_{tp},\!\mathbf{p}_{tp}\right)/\tau\right)}{\sum_{{tp}^{\prime}\neq {tp}}\exp\left(\operatorname{sim}\left(\mathbf{p}_{tp}, \mathbf{p}_{{tp}^{\prime}}\right)/\tau\right)},
% \vspace{-5pt}
\end{equation}
}
% For the sake of brevity, we will drop the dependence of the loss function on temperature hyperparameter $\tau$ in the following. 
where $T$ is textual features, $\operatorname{sim}(\cdot,\cdot)$ denotes the cosine similarity of the embedded features, and $\tau$ is the temperature:
This process constructs separable semantic anchors from the clean and stable textual space, providing reliable guidance for subsequent visual partitioning.

\noindent\textbf{Vision space partition.}
Similar to the textual modality, we construct base, novel, and background prototypes in the visual embedding space to establish partition-level geometry. However, unlike text embeddings, visual representations suffer from missing annotations for novel categories and semantic overlap with background regions, which makes direct prototype learning non-trivial.

\textit{\textbf{1) Prototype construction.}}
Base prototypes are computed by averaging region embeddings from ground-truth boxes and their matched positive proposals, while background prototypes are formed from negative proposals defined with respect to ground-truth annotations. The challenge arises for novel categories that lack bounding-box annotations and thus reliable proposals. Existing pseudo-labeling strategies, including mining novel-category samples with vision–language models or external caption datasets, often introduce noisy supervision and inaccurate region–word alignment \cite{52_cckt}.

To obtain cleaner visual anchors for the novel partition, we propose a generative pseudo-labeling strategy termed structure prompt (SP). SP leverages LLM-expanded (GPT-4o \cite{76_gpt4o}) descriptions to guide a diffusion model (SD-3.5 \cite{77_sd3.5}) in synthesizing representative novel images, incorporating both intrinsic object attributes (\textit{e.g.}, shape, color, material) and contextual cues (\textit{e.g.}, scene context and co-occurring objects) \cite{50_cognition}.
We extract multi-scale features of the generated images by interpolating activation maps from multiple layers of ViT-L/14 \cite{33_clipself} and fuse them via a Feature Pyramid Network (FPN), relying solely on the inter-class semantic structure of generated images rather than their visual fidelity.

% Different from embedding-level proxy synthesis methods such as ProxyDet~\cite{16_proxydet}, which generate proxy-novel classes by class-wise mixup between base-class embeddings, SP constructs visual anchors from explicit novel-category semantics. It organizes appearance, context, and co-occurrence cues to guide diffusion-based image synthesis, and the extracted multi-scale features are then used as novel-category visual prototypes in PR-OVD. Thus, SP is a structured visual-anchor construction mechanism that supports representation-space partitioning.

\textit{\textbf{2) Asymmetric prototype learning.}}
Another critical issue arises from the semantic overlap between novel objects and background regions. Since novel instances lack annotations, many of their regions are \textit{implicitly treated as background} during training. As a result, enforcing negative contrast between novel and background prototypes may contaminate the novel partition and blur boundaries, as evidenced in Tab.\ref{table:abcvd}.

To mitigate this effect, we adopt an asymmetric prototype learning strategy in partition-level visual prototype learning \cite{46_triple}, where contrastive optimization is performed only on reliable base–novel and base–background relationships, while the novel–background relationship is left unconstrained due to the lack of reliable novel-category annotations. For each visual partition \(vp\), we collect the set of region embeddings \(R_{vp}\) belonging to that partition and compute its prototype representation as $\mathbf{p}_{vp}=\frac{1}{\left|R_{vp}\right|} \sum_{v \in R_{vp}} f_{\operatorname{visual}}(v)$. The partition-level prototype learning objective is defined as:
{
\small
\begin{equation}
\mathcal{L}_{\operatorname{cv}}(I,I^+,I^-)=-\sum_{vp \in \{\text{base}\}}
\log \frac{
\exp(\mathrm{sim}(\mathbf{p}_{vp}, \mathbf{p}_{vp})/\tau)
}{
\sum_{x}\exp(\mathrm{sim}(\mathbf{p}_{vp}, \mathbf{p}_x)/\tau)
},
\label{equ:i2i}
\end{equation}
}
\noindent where $I$ is image features and $x \in \{\text{novel},\text{bg}\}$. 
Notably, we do not enforce explicit negative contrast between novel and background instances. This intentional design mitigates the risk of semantic confusion caused by latent novel instances within background regions, thereby reinforcing the reliability of category-aware semantic modeling.

In the coarse partition process, category-level distinctions within each partition are deliberately ignored, focusing solely on establishing clear boundaries to prevent cross-partition confusion and provide a stable foundation for subsequent region-level and fine-grained alignment, as illustrated in Fig.\ref{fig:method}.

\subsection{Category-level Refinement Module}
While the coarse partition module establishes global separation, giving novel categories dedicated regions during training, it still lacks inter-class constraints. Therefore, we perform category-level refinement to sharpen inter-class boundaries and correct cross-modal misalignment within the partitioned geometry. Moreover, since CLIP \cite{27_clip} pretraining relies on image-level supervision using the \texttt{CLS} token and lacks region-level alignment, we introduce a bidirectional, multi-modal region-level learning strategy to force discriminative alignment between local visual features and their corresponding text embeddings, enhancing fine-grained semantic consistency.

In particular, visual embeddings include supervised base/background proposals as well as structure prompt–generated features of novel classes, while textual embeddings are derived from an open vocabulary.
Following the InfoNCE loss \cite{34_infonce}, we formulate image-to-text and text-to-image matching objective by:
% \vspace{-5pt}
{
\small
% \begin{equation}
% \mathcal{L}_{\operatorname{v2t}}\!\left(I,\! T^{+},\! T^{-}\right)\!=\!-\!\mathbb{E}_{p(I, T)}\!\left[\!\log \!\frac{\exp \left(\operatorname{sim}\!\left(I,\!T^{+}\right) \!/\! \tau\right)}{\sum_{k=1}^K \exp \left(\operatorname{sim}\!\left(I,\!T_k^{-}\right)\!/\! \tau\right)}\!\right] \label{multi_v2t} 
% \end{equation}
\begin{equation}
\mathcal{L}_{\operatorname{v2t}}\left(I, T^{+}, T^{-}\right)
\!=-\!\mathbb{E}_{p(I\!,T\!)}\!\left[\!\log \!\frac{
\exp\!\left(\operatorname{sim}\!\left(I,\!T^{+}\right)/ \tau\right)
}{
\sum_{k=1}^K\! \exp\!\left(\operatorname{sim}\!\left(I,\!T_k^{-}\right)/ \tau\right)
}\!\right]
\label{multi_v2t}
\end{equation}
}
{
\small
% \begin{equation}
% \mathcal{L}_{\operatorname{t2v}}\left(T, I^{+}, I^{-}\right) =-\mathbb{E}_{p(I, T)}\left[\log \frac{\exp \left(\operatorname{sim}\left(T, I^{+}\right) / \tau\right)}{\sum_{k=1}^K \exp \left(\operatorname{sim}\left(T, I_k^{-}\right) / \tau\right)}\right] \label{multi_t2v}
% \end{equation}
\begin{equation}
\mathcal{L}_{\operatorname{t2v}}\left(T, I^{+}, I^{-}\right) 
=-\!\mathbb{E}_{p(I\!,T\!)}\!\left[\!\log \!\frac{
\exp\!\left(\operatorname{sim}\!\left(T,\!I^{+}\right)/ \tau\right)
}{
\sum_{k=1}^K\! \exp\!\left(\operatorname{sim}\!\left(T,\!I_k^{-}\right)/ \tau\right)
}\!\right] 
\label{multi_t2v}
\end{equation}
}

\noindent where $T^{-}$ and $I^{-}$ denote sets of negative text and image samples negative text and image samples not aligned with the category of $T$ and $I$.

Inspired by human coarse-to-fine perception \cite{72_human1}, our PR-OVD organizes the embedding space into global partitions, explicitly reserving regions for novel categories and mitigating semantic collapse. Building on this structured geometry, it performs category-level refinement to sharpen inter-class boundaries, consolidate intra-class clusters, and enforce region-level alignment between visual features and text embeddings. This unified, partition-aware framework preserves novel-category space while enhancing cross-modal discriminability. All losses from the preceding modules are integrated to form the complete partition-and-refine learning objective:  
{
% \small
\begin{equation}\label{equ:contras}
\begin{aligned}
L_{PR} 
= \frac{1}{4} \Big[
    &L_{\operatorname{ct}}\left(T, T^{+}, T^{-}\right) 
    + L_{\operatorname{v2t}}\left(I, T^{+}, T^{-}\right) \\
    &+ L_{\operatorname{t2v}}\left(T, I^{+}, I^{-}\right) 
    + L_{\operatorname{cv}}\left(I, I^{+}, I^{-}\right)
\Big]
\end{aligned}
\end{equation}
}
% {
% \small
% \begin{equation}\label{equ:contras}
% L_{PR} 
% = \frac{1}{4} \Big[
%     L_{\operatorname{ct}}\left(T, T^{+}, T^{-}\right) 
%     + L_{\operatorname{v2t}}\left(I, T^{+}, T^{-}\right)
%     + L_{\operatorname{t2v}}\left(T, I^{+}, I^{-}\right) 
%     + L_{\operatorname{cv}}\left(I, I^{+}, I^{-}\right)
% \Big]
% \end{equation}
% }

% \noindent More details of our PR-OVD are provided in the supplementary material.

\subsection{Training and Inference}
\textbf{Training phase.} We adopt a standard two-stage object detection framework \cite{33_clipself}, where the total loss function is formulated as:
{
\begin{equation}
    {L_{\operatorname{total}} = L_{\operatorname{PR}}} + L_{\operatorname{rpn}} + L_{\operatorname{reg}} + L_{\operatorname{cls}}
\end{equation}
}
Here, $\boldsymbol{L_{PR}}$ is our proposed partition-and-refine objective, explicitly involving novel and background categories in training and enforcing inter-class constraints to alleviate semantic collapse and calibration imbalance. $\boldsymbol{L_{\operatorname{rpn}}}$ denotes the Region Proposal Network (RPN) loss, typically implemented as binary cross-entropy (BCE) loss to distinguish foreground from background proposals. $\boldsymbol{L_{\operatorname{reg}}}$ represents the bounding box regression loss, computed using the standard Smooth L1 loss for precise localization. $\boldsymbol{L_{\operatorname{cls}}}$ computes the similarity between region features of base categories and textual embeddings, followed by softmax normalization to obtain categorical probabilities.

\noindent\textbf{Inference phase.} The detector performs class prediction by computing similarity scores between regional and textual embeddings of all candidate categories. This mechanism inherently equips the model with open-vocabulary capabilities, enabling seamless extension to novel categories that are unseen during training.

\begin{table*}[t]
\centering
% \vspace{-6pt}
\setlength{\tabcolsep}{8pt}
\renewcommand{\arraystretch}{1.0}
\captionof{table}{\textbf{Comparisons with SoTA OVD methods on OV-COCO \cite{10_ovrcnn} and OV-LVIS \cite{36_vild}.} 
Methods are categorized and compared according to the taxonomy described in Sec.\ref{sec:rl}. 
Bolded values indicate the best performance. Following prior works, we focus on the method’s performance on novel categories.}
\resizebox{\linewidth}{!}{
\begin{tabular}{l|l|l|l|>{\columncolor[HTML]{EAEAFF}}c>{\color[gray]{0.6}}c>{\color[gray]{0.6}}cc>{\columncolor[HTML]{EAEAFF}}c>{\color[gray]{0.6}}c>{\color[gray]{0.6}}c>{\color[gray]{0.6}}c}
\toprule
\multirow{2}{*}{\textbf{Category}}     & \multirow{2}{*}{\textbf{Methods}} & \multirow{2}{*}{\textbf{Venue}} & \multirow{2}{*}{\textbf{Backbone}} & \multicolumn{3}{c}{\textbf{OV-COCO}}                     & \multicolumn{5}{c}{\textbf{OV-LVIS}}        \\ \cmidrule(l){5-7}\cmidrule(l){9-12} 
&   & &  & \textbf{$\text{AP}_\text{50}^\text{novel}$} & $\text{AP}_\text{50}^\text{base}$ & $\text{AP}_\text{50}^\text{all}$ & &\textbf{$\text{mAP}_\text{novel}$} & $\text{mAP}_\text{c}$ & $\text{mAP}_\text{f}$ & $\text{mAP}_\text{all}$ \\ \midrule

\multirow{8}{*}{\shortstack[l]{\textbf{Pseudo} \\\textbf{Label} \\\textbf{related}}} & OV-RCNN \cite{10_ovrcnn}   & CVPR-21   &   RN50-C4 (24M)        & 17.5              & 41.0             & 34.9        &    & -             & -         &  -                                 & -           \\
% & PB-OVD \cite{12_pbovd}                             & 30.8              & 46.1             & 42.1 &  & -             & -         &      -                          & -           \\
& CORA \cite{23_cora}       & CVPR-23  & RN50×4 (87M)     & 35.1              & 35.5            & 35.4      &      & 22.2             & -     &  -        & -      \\
& VLDet \cite{49_vldet}      & ICLR-23   & RN50-C4 (24M)      & 32.0              & 50.6            & 45.8      &      & 26.3             & 39.4     &  41.9        & 38.1      \\
& SAS-Det \cite{14_sasdet}    & CVPR-24  & RN50-C4 (24M)      & 37.4       & 58.5
   & 53.0       &     & 20.9             & 26.1        & 31.6    & 27.4           \\
 & RALF \cite{15_ralf}         &CVPR-24   & RN50 (24M)  & 41.3              & 54.3           & 50.9      &      & 21.9             & 26.2         & 29.1         & 26.6        \\
& LBP \cite{48_lbp}      & CVPR-24  & RN50 (24M)  & 37.8          & 58.7        & 53.2  &      & 24.1         & 29.5         & 32.8       & 29.9          \\
& ProxyDet \cite{16_proxydet}  & AAAI-24  & RN50 (24M)   & 30.4          & 52.6        & 46.8  &      & 18.9        & -         &  -   & 30.1          \\
& CCKT-Det \cite{52_cckt}    & ICLR-25  & SwinB (88M) & 41.9          & -        & 40.9  &      & -        & -         &  -   & -          \\\midrule

\multirow{6}{*}{\shortstack[l]{\textbf{CLIP} \\\textbf{related}}}          & RegionCLIP \cite{11_regionclip}   & CVPR-22 & RN50×4 (87M)  & 39.3   & 61.6  & 55.7  &              & 22.0    & 32.1          & 36.9   & 32.3              \\
 & F-VLM \cite{38_fvlm} &ICLR-23  & RN50-FPN (24M)  & 28.0     & -      & 39.6       &         & 18.6            & -     & -      & 24.2      \\
& RO-ViT \cite{43_rovit} &CVPR-23 & ViT-L/16 (307M)     & 33.0     & -      & 47.7       &     & 32.1                & -             & -            & 34.0               \\ 
& CFM-ViT \cite{44_cfmvit}  &ICCV-23 & ViT-L/16 (307M)  & 34.1   & -     & 46.0   &  & 33.9                 & -             &  -       & 36.6              \\
& CLIPSelf \cite{33_clipself} &ICLR-24  & ViT-B/16 (86M)   & 44.3     & 64.1   & 59.0     &       &34.9     & 34.6      &  35.6    & 35.1              \\
& SIC-CAD \cite{47_siccad} &AAAI-24 & RN50-FPN (24M) & 43.3   & 45.7    & 45.1   &         & 21.9                & 32.0            & 35.5         & -               \\
& UOVN \cite{81_uovn}  &TMM-24  & Swin-L (198M)  & -   &  -    & -   &         & 29.1                & 41.2            & 43.6        & 40.3               \\
& CAKE \cite{51_cake}  &AAAI-25  & RN50 (24M)  & 41.8   & 60.6    & 55.7   &         & 25.0                & 34.8            & 38.4        & 34.9               \\
& Det-Agent \cite{79_detagent}  &TMM-26  & RN50×4 (87M)  & 39.6   &  53.9    & 50.4   &         & 21.8                & -            & -        & -               \\
\midrule
{\textbf{Proposed}} & \textbf{PR-OVD} &--  & ViT-B/16 (86M)  & \textbf{47.4}          & 63.3         & 59.1    &    & \textbf{37.0}         & 34.0         & 35.3    & 35.1           \\\bottomrule
\end{tabular}%
}
\label{tab:comparison}
\vspace{-7pt}
\end{table*}

\section{Experiments}
\label{sec:exper}

\subsection{Experimental Setup}
% \paragraph{Benchmark Details.}
\textbf{Benchmark details.}
We evaluate our method on two widely used benchmarks for OVD: OV-COCO and OV-LVIS. OV-COCO, introduced by OV-RCNN \cite{10_ovrcnn}, divides the 65 object categories in COCO \cite{40_coco} into 48 base and 17 novel categories. 
% excluding 15 categories that do not have synonym sets in WordNet. 
The performance is primarily assessed using box Average Precision (AP) at an IoU threshold of 0.50. OV-LVIS, proposed by ViLD \cite{36_vild}, contains 1,203 long-tailed categories from LVIS v1.0 \cite{41_lvis}, with 337 rare categories designated as novel classes and the rest as common or frequent.
% contains 1,203 categories in a long-tail distribution from the LVIS v1.0 \cite{41_lvis} dataset and selects 337 rare categories as novel classes, with the remaining categories treated as common or frequent. 
The evaluation focuses on the mean mask AP for the novel (rare) categories.

We further evaluate our PR-OVD on other open-vocabulary tasks, including semantic segmentation and multi-label classification. For segmentation, following MaskAdapter \cite{59_maskadapter}, we train on COCO-Stuff \cite{64_cocostuff} (118K images, 171 categories) and evaluate on ADE20K (A-150, A-847) \cite{67_ade} and Pascal-Context (PC-59, PC-459) \cite{66_pascalcontext} using mean Intersection-over-Union (mIoU). For classification, we adopt the RAM \cite{63_ram} protocol and validate on NUS-WIDE \cite{65_nuswide} (925 seen, 81 unseen categories), reporting top-3 precision (P@3), F1@3, and mean Average Precision (mAP).

\noindent\textbf{Implementation details.}
Following prior work \cite{33_clipself,38_fvlm}, we adopt Mask R-CNN \cite{6_mrcnn} with a Feature Pyramid Network (FPN) \cite{42_fpn} as the detection head and ViT-L/14 \cite{43_rovit,44_cfmvit} as the visual backbone. We freeze the CLIP encoder while training only the Mask R-CNN parameters. PR-OVD utilizes the 6th, 10th, 14th, and 23rd layers of ViT-L/14 to construct visual semantics, corresponding to relative image scales of 1/3.5, 1/7, 1/14, and 1/28, respectively \cite{33_clipself}. These components follow the standard detector configuration for ViT-based object detection and are not introduced by PR-OVD. As a detector-agnostic framework, PR-OVD can be seamlessly integrated into different detector architectures without introducing additional structural modifications. The detector is trained on 8 NVIDIA GeForce RTX 3090 GPU with a batch size of 8 per GPU. We train for 3 epochs on OV-COCO and 48 epochs on OV-LVIS, using a fixed input resolution of 896×896. The AdamW optimizer is used with a learning rate of 1e-4 and a weight decay of 0.1. 
% After color normalization, we pad the bottom and right sides of the image with zeros to ensure compatibility with models that require square inputs.

\begin{figure*}[t]
\centering
\includegraphics[width=0.7\textwidth]{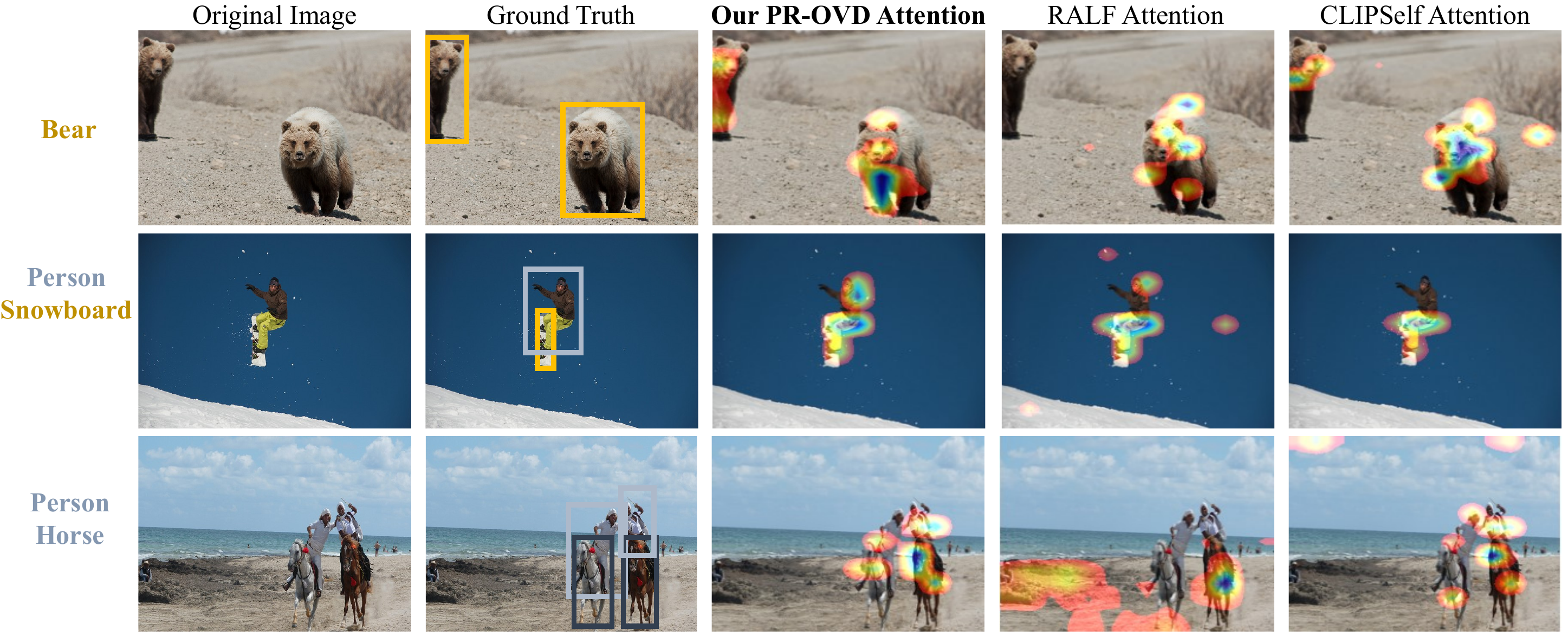}
\caption{\textbf{Qualitative comparison of attention maps on \textcolor[HTML]{90A1B7}{\textbf{base}} and \textcolor[HTML]{BF9000}{\textbf{novel}} categories with SoTAs}. Compared with CLIPself \cite{33_clipself} and RALF \cite{15_ralf}, PR-OVD produces more precise attention, particularly for \textcolor[HTML]{BF9000}{\textbf{novel}} categories.}
% Red and blue bounding boxes represent novel and base categories, respectively.} 
\label{fig:attn}
\vspace{-10pt}
\end{figure*}

\subsection{Main results}

\noindent\textbf{Quantitative comparison with SoTAs.} Tab.\ref{tab:comparison} presents the results of our method against SoTA methods on the OV-COCO and OV-LVIS benchmarks. 
While the former also use CLIP, they do not explicitly tailor their designs to it.
Compared with the second-best method, CLIPSelf \cite{33_clipself}, our method achieves an average \textbf{6.5\%} proportional improvement across the two datasets, while maintaining competitive results on both `base' and `all' categories. Specifically, on OV-COCO, PR-OVD achieves $\text{AP}_\text{50}^\text{novel}$ of 47.4, surpassing the best-performing pseudo-label method CCKT-Det \cite{52_cckt} by a significant margin of \textbf{5.5} $\text{AP}_\text{50}^\text{novel}$.
For consistency, we only compare with the CCKT-Det variant without inference-time enhancements.
On the more challenging OV-LVIS, PR-OVD shows strong generalization to novel categories, surpassing the best pseudo-label method SAS-Det \cite{14_sasdet} by a substantial margin of \textbf{7.9} mAP$_\text{novel}$. These results validate the effectiveness of PR-OVD in reserving dedicated space for novel categories during training and reinforcing inter-class constraints to improve performance. 
On OV-LVIS, although the improvement on the “All” metric is limited, PR-OVD achieves consistent gains on novel categories. This is mainly because the long-tailed distribution makes the overall metric dominated by frequent seen categories, reducing its sensitivity to rare novel-category improvements. 
% These results demonstrate that PR-OVD effectively reallocates representation capacity toward unseen categories while maintaining competitive seen-category performance.

\noindent\textbf{Qualitative comparison with SoTAs.}
We compare our method with CLIPSelf \cite{33_clipself} and RALF \cite{15_ralf} by visualizing the top 90\% of attention from each model’s final layer overlaid on the original images.
As shown in Fig.\ref{fig:attn}, both CLIPSelf and RALF often exhibit coarse and unstable attention, often distracted by spurious background activations and missing category-relevant cues. For example, RALF largely ignores a partially visible novel-class bear (\textit{top row, fourth column}) while focusing on irrelevant ground regions (\textit{third row, fourth column}).
In contrast, PR-OVD jointly constructs a global partitioned space, reserving regions for novel and background categories and enforcing inter-class constraints, thereby suppresses spurious, object-irrelevant attention while preserving fine-grained, object-centered signals.

\subsection{Ablation Study}
\label{sec:ablation}
\begin{figure}[t]
\centering
% \vspace{-22pt}
\includegraphics[width=0.8\columnwidth]{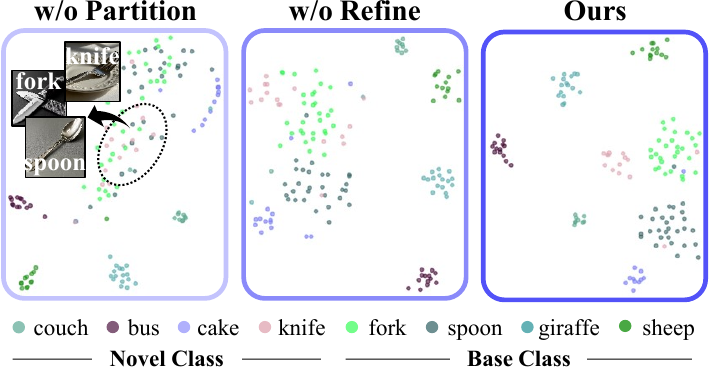}
\caption{\textbf{Qualitative analysis of module contributions} on the OV-COCO benchmark using t-SNE feature visualization.} 
\label{fig:visual_level}
% \vspace{-20pt}
\end{figure}

In this section, we evaluate the contributions of each component within our PR-OVD. We first assess the impact of two main modules: coarse partition and category-level refinement.(Tab.\ref{table:ablevel} and Fig.\ref{fig:visual_level}). We then investigate the design choices within the coarse partition process (Tab.\ref{table:abconcept}), including the effectiveness of asymmetric prototype learning in the visual space (Tab.\ref{table:abcvd}) and the role of our Structured Prompt (SP) for novel-class prototype construction (Tab.\ref{table:abconcept}).

\begin{table}[t]
\centering
\setlength{\tabcolsep}{13pt}
\captionof{table}{\textbf{Quantitative analysis of module contributions} on the OV-COCO \cite{10_ovrcnn} benchmark.} 
\resizebox{\columnwidth}{!}{
\begin{tabular}{cc>{\columncolor[HTML]{EAEAFF}}ccc}
% \hline
\toprule
\multirow{2}{*}{\begin{tabular}[c]{@{}c@{}}Coarse\\ partition\end{tabular}} & \multirow{2}{*}{\begin{tabular}[c]{@{}c@{}}Category-level\\ refine\end{tabular}} & \multicolumn{3}{c}{OV-COCO} \\
\cmidrule(lr){3-5}
& & $\text{AP}_\text{50}^\text{novel}$  & $\text{AP}_\text{50}^\text{base}$ & $\text{AP}_\text{50}^\text{all}$  \\ \midrule
\ding{55}              & \ding{51}              & 44.8  & 62.2 & 57.3 \\
\ding{51}             & \ding{55}               & 45.9  & 61.5 & 58.0 \\
\ding{51}            & \ding{51}              & \textbf{47.4}  & \textbf{63.3} & \textbf{59.1} \\ \bottomrule
\end{tabular}%
}
\label{table:ablevel}
% \vspace{-25pt}
\end{table}

\noindent\textbf{Effectiveness of modules.}
To assess the performance contribution of different modules in PR-OVD, we conduct quantitative analysis as illustrated in Tab.\ref{table:ablevel}. Each module contributes from a distinct perspective, and their combination yields the strongest results, highlighting the complementary roles of global partitioning and category-level refinement.

To gain clearer insight into the effectiveness of each module, we perform a thorough qualitative analysis by employing t-SNE to show the feature distributions of a subset of base and novel categories, as shown in Fig.\ref{fig:visual_level}. 
With category-level refinement alone, some categories are well separated, but semantically similar base and novel classes (\textit{e.g.}, novel “knife” vs. base “fork” and “spoon”) tend to be entangled due to the absence of a global partition geometry. 

Conversely, with only coarse partition, base and novel classes are separated at a global level, yet intra-class features are less compact and inter-class margins remain limited. 
When both component are combined, all categories are clearly distinguishable. Specifically, the coarse partition establishes a global semantic structure that reserves dedicated space during training for categories unseen in training but required at test time, preventing them from being absorbed by dominant base classes. The category-level refinement then sharpens intra-partition structures by enforcing inter-class constraints and enhancing decision boundaries. Their synergy provides both global structural guidance and fine-grained discriminability, leading to substantial performance gains.

\begin{table}[t]
    \centering
    \setlength{\tabcolsep}{10pt}
    \captionof{table}{\textbf{Ablation of coarse partition module architecture} on the OV-COCO benchmark.} 
    \label{table:abconcept}
    \resizebox{\columnwidth}{!}{ % 撑满这 0.48 的宽度
        \begin{tabular}{cc>{\columncolor[HTML]{EAEAFF}}ccc}
            \toprule
            \multirow{2}{*}{\begin{tabular}[c]{@{}c@{}}Language space\\ partition\end{tabular}} & \multirow{2}{*}{\begin{tabular}[c]{@{}c@{}}Vision space\\ partition\end{tabular}} & \multicolumn{3}{c}{OV-COCO} \\ \cmidrule(lr){3-5} 
             & & $\text{AP}_\text{50}^\text{novel}$ & $\text{AP}_\text{50}^\text{base}$ & $\text{AP}_\text{50}^\text{all}$ \\ \midrule
            \ding{55} & \ding{55} & 44.8 & 62.2 & 57.3 \\
            \ding{55} & \ding{51} & 46.0 & 61.8 & 57.7 \\
            \ding{51} & \ding{55} & \underline{46.5} & \underline{62.4} & \underline{58.1} \\
            \ding{51} & \ding{51} & \textbf{47.4} & \textbf{63.3} & \textbf{59.1} \\
            \bottomrule
        \end{tabular}
    }
\end{table}

\begin{table}[t]
    \centering
    \setlength{\tabcolsep}{12pt}
    \captionof{table}{\textbf{Ablation of asymmetric prototype learning strategy} on the OV-COCO benchmark.} 
    \label{table:abcvd}
    \resizebox{\columnwidth}{!}{
        \begin{tabular}{ccc>{\columncolor[HTML]{EAEAFF}}ccc}
            \toprule
            \multirow{2}{*}{BG-N} & \multirow{2}{*}{BG-B} & \multirow{2}{*}{N-B} & \multicolumn{3}{c}{OV-COCO} \\ \cmidrule(lr){4-6} 
            & & & $\text{AP}_\text{50}^\text{novel}$ & $\text{AP}_\text{50}^\text{base}$ & $\text{AP}_\text{50}^\text{all}$ \\ \midrule
            \ding{51} & \ding{51} & \ding{51} & 45.6 & 61.7 & 57.5 \\
            \ding{55} & \ding{51} & \ding{55} & 46.2 & 62.6 & 58.2 \\
            \ding{55} & \ding{55} & \ding{51} & \underline{46.4} & \textbf{63.4} & \underline{58.9} \\ 
            \ding{55} & \ding{51} & \ding{51} & \textbf{47.4} & \underline{63.3} & \textbf{59.1} \\
            \bottomrule
        \end{tabular}
    }
\end{table}

\noindent\textbf{Effectiveness of partition design. }
To validate the architecture of the coarse partition module, we keep the category-level refinement process fixed and evaluate different configurations, as outlined in Tab.\ref{table:abconcept}. The results reveal that building a linguistic partition via language space significantly boosts performance, as it provides a clearer global semantic scaffold for representation learning. When visual space is further integrated, the cooperation between modalities brings an additional gain, showing that global semantic coherence is best achieved through joint language–vision structuring.

\begin{table}[t]
\centering
% \vspace{-34pt}
\setlength{\tabcolsep}{5pt}
\captionof{table}{\textbf{Ablation study on the impact of SP in PR-OVD} on the OV-COCO and OV-LVIS benchmarks.} 
\resizebox{\columnwidth}{!}{%
\begin{tabular}{@{}c>{\columncolor[HTML]{EAEAFF}}cccl>{\columncolor[HTML]{EAEAFF}}cccc@{}}
\toprule
\multirow{2}{*}{\begin{tabular}[c]{@{}c@{}}Structure \\ Prompt\end{tabular}} & \multicolumn{3}{c}{OV-COCO} &  & \multicolumn{4}{c}{OV-LVIS} \\ \cmidrule(lr){2-4} \cmidrule(l){6-9} 
& $\text{AP}_\text{50}^\text{novel}$ & $\text{AP}_\text{50}^\text{base}$ & $\text{AP}_\text{50}^\text{all}$    &  & $\text{mAP}_\text{novel}$ & $\text{mAP}_\text{c}$ & $\text{mAP}_\text{f}$ & $\text{mAP}_\text{all}$  \\ \midrule
\ding{55}                                & 46.2     & 62.5    & 57.8   &  & 36.2   & 34.0 & 35.2 & 35.0 \\
\ding{51}                                & \textbf{47.4}     & 63.3    & 59.1   &  & \textbf{37.0}   & 34.0 & 35.3 & 35.1 \\ \bottomrule
\end{tabular}%
}
\label{table:sp}
% \vspace{-25pt}
\end{table}

\begin{figure}[t]
\centering
% \vspace{-20pt}
\includegraphics[width=0.8\columnwidth]{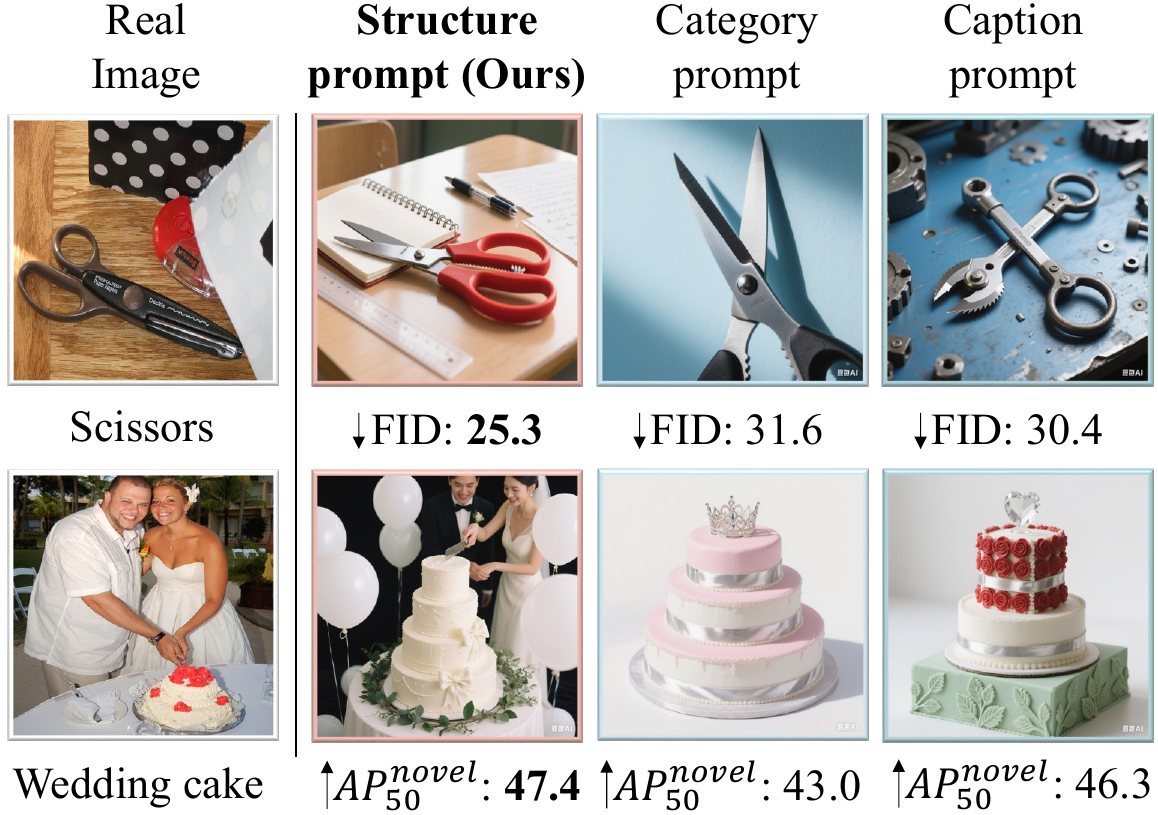}
\caption{\textbf{Qualitative and quantitative evaluation of novel-category image generation strategy.} FID measures the distribution gap between generated and real images, with lower values indicating better quality and realism. $\text{AP}_\text{50}^\text{novel}$ reports the detection performance on novel categories in OV-COCO.} 
\label{fig:visual_sp}
% \vspace{-10pt}
\end{figure}

% \paragraph{Effectiveness of Embedding Separation Strategies.}
\noindent\textbf{Effectiveness of asymmetric prototype learning strategy.}
To investigate the impact of imposing constraints within the visual space, focusing on negative contrasts among base, novel, and background visual partition. As shown in Tab.\ref{table:abcvd}, enforcing full triplet contrast degrades performance due to the introduced noise from latent novel instances mislabeled as background. In contrast, avoiding explicit background–novel contrast (`Bg-N') consistently improves results, while retaining only `Bg–B’ and `N–B’ yields the best outcome. These findings confirm that avoiding direct background–novel comparison, yet maintaining their separation from base categories, leads to more stable and effective open-vocabulary learning.

\begin{table}[t]
\centering
\setlength{\tabcolsep}{14pt}
\captionof{table}{\textbf{Ablation study on different LLMs for structure prompt generation} on the OV-COCO benchmark.}
\resizebox{\columnwidth}{!}{
\begin{tabular}{l>{\columncolor[HTML]{EAEAFF}}lll}
\toprule
\multicolumn{1}{l}{\multirow{2}{*}{LLM Type}} & \multicolumn{3}{c}{OV-COCO} \\ \cmidrule(l){2-4} 
\multicolumn{1}{c}{}                          & $\text{AP}_\text{50}^\text{novel}$ & $\text{AP}_\text{50}^\text{base}$ & $\text{AP}_\text{50}^\text{all}$   \\ \midrule
GPT-4o \cite{76_gpt4o}                        & 47.4     & 63.3           & \textbf{59.1}   \\
Qwen2.5-7B-Instruct \cite{86_qwen}             & 47.2     & \textbf{63.4}  & \textbf{59.1}   \\
Llama-3.1-8B-Instruct \cite{87_llama}          & \textbf{47.6} & 63.1      & 59.0   \\ 
\bottomrule
\end{tabular}%
}
\label{tab:llm}
\end{table}

\begin{figure}[t]
\centering
% \vspace{-20pt}
\includegraphics[width=0.75\columnwidth]{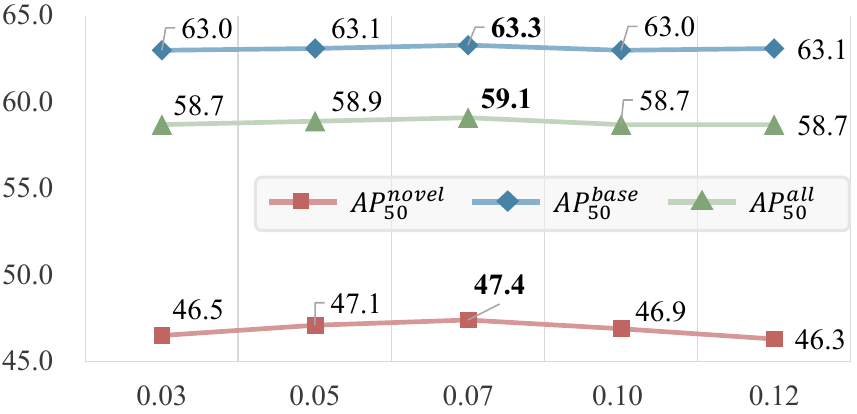}
\caption{\textbf{Sensitivity analysis of the temperature hyperparameter $\tau$.} We report AP$_{50}^{novel}$, AP$_{50}^{base}$, and AP$_{50}^{all}$ on OV-COCO under different values of $\tau$ used in Eq.~(\ref{equ:t2t})--Eq.~(\ref{multi_t2v}).}
\label{fig:ab_temp}
% \vspace{-10pt}
\end{figure}

\begin{figure}[t]
\centering
% \vspace{-20pt}
\includegraphics[width=0.75\columnwidth]{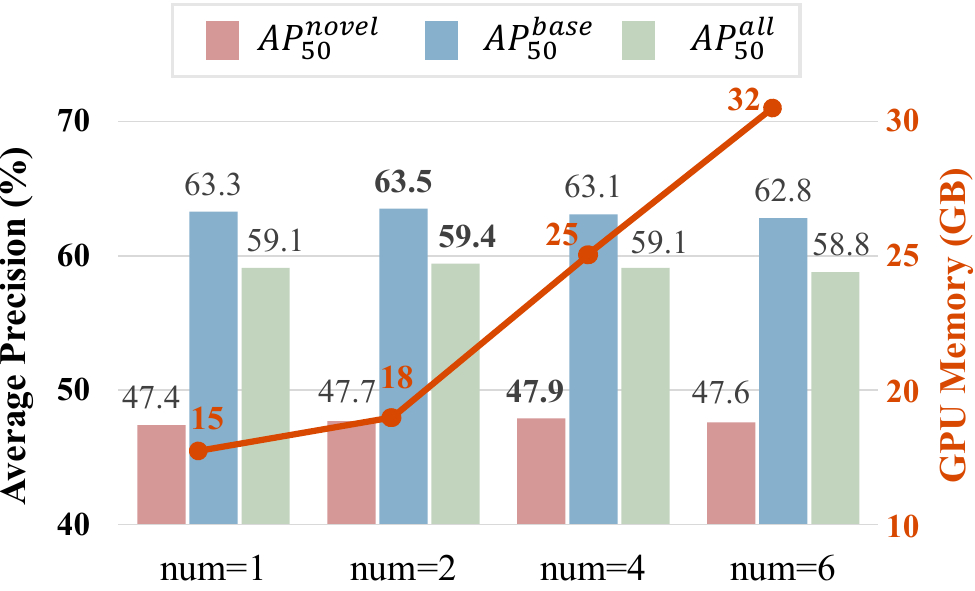}
\caption{\textbf{Sensitivity analysis of the number of synthetic images.} We evaluate different numbers of synthetic images per novel category on the OV-COCO benchmark. Bar charts report AP$_{50}^{novel}$, AP$_{50}^{base}$, and AP$_{50}^{all}$, while the line curve indicates GPU memory usage.}
\label{fig:ab_num}
% \vspace{-10pt}
\end{figure}

% \paragraph{Effectiveness of Structure Prompt Design.}
\noindent\textbf{Effectiveness of structure prompt design.}
To validate the contribution of structure prompt which generating novel-category samples within the overall PR-OVD framework, we keep the other process fixed and report quantitative results in Tab.\ref{table:sp}. Incorporating SP introduces multi-scale novel visual anchors, which consistently boosting performance on both base and novel categories, demonstrating its positive impact on the model’s representation quality.

To evaluate SP as a prompting strategy for generating visual features of novel categories, we compare it with two common alternatives: 
\noindent(1) raw category names (\textit{e.g.}, ``scissors") and (2) caption-derived descriptions (\textit{e.g.}, ``a tool having two crossed pivoting blades with looped handles"). Qualitative and quantitative comparisons on OV-COCO are shown in Fig.\ref{fig:visual_sp}. SP-generated images exhibit stronger semantic alignment with real-world scenes, as reflected in lower Fréchet Inception Distance (FID) scores. This advantage is especially evident in fine-grained categories (\textit{e.g.}, ``wedding cake" vs. ``birthday cake"), where SP-guided generation captures critical contextual distinctions. Furthermore, our SP consistently leads to higher performance in novel category ($\text{AP}_\text{50}^\text{novel}$), validating the importance of structure prompt design for robust OVD.

To further improve reproducibility and evaluate the robustness of our method to the choice of LLMs, we replace the closed-source GPT-4o with two open-weight LLMs (Qwen2.5-7B-Instruct~\cite{86_qwen} and Llama-3.1-8B-Instruct~\cite{87_llama}), while keeping the remaining pipeline unchanged. As shown in Tab.~\ref{tab:llm}, both open-weight LLMs achieve comparable performance to GPT-4o on OV-COCO, indicating that PR-OVD does not rely on a specific proprietary LLM. This is because the LLM in our framework is mainly used to organize category semantics into structured descriptions, including appearance, context, and co-occurrence cues, rather than to provide direct detection supervision. Furthermore, we fix GPT-4o as the LLM and evaluate two different diffusion models, i.e., SD-1.5 and SD-3.5. The performance gap between them is within approximately 0.3 AP, suggesting that our method is also insensitive to the specific image generator. These results demonstrate that the effectiveness of PR-OVD mainly stems from the proposed structure prompt and partition-refine learning framework, rather than from a particular proprietary LLM or generative model.

\noindent
\textbf{Sensitivity of  temperature hyperparameter $\tau$.}
We further investigate the influence of the temperature hyperparameter $\tau$, which is consistently used in Eq.~(\ref{equ:t2t})--Eq.~(\ref{multi_t2v}) to control the sharpness of contrastive optimization. As shown in Fig.~\ref{fig:ab_temp}, PR-OVD exhibits stable performance across a wide range of temperature values, demonstrating that our framework is not overly sensitive to this hyperparameter. Specifically, varying $\tau$ from 0.03 to 0.12 only leads to moderate fluctuations. Among these settings, $\tau=0.07$ achieves the best overall performance, obtaining 47.4 AP$_{50}^{novel}$, 63.3 AP$_{50}^{base}$, and 59.1 AP$_{50}^{all}$. A smaller $\tau$ makes the contrastive distribution sharper and may slightly increase the sensitivity to hard negatives and noisy synthetic visual prototypes, while a larger $\tau$ smooths the distribution and mildly weakens the separation between different partitions. These results indicate that $\tau=0.07$ provides a favorable balance between inter-partition separation and fine-grained semantic preservation, and we therefore adopt it as the default setting.

\noindent
\textbf{Sensitivity to the number of synthetic images.}
To investigate the optimal number of synthetic images per novel category for stabilizing the visual prototype $p_{vp}$, we evaluate different settings on OV-COCO and analyze the trade-off between detection performance and GPU memory usage, as shown in Fig.~\ref{fig:ab_num}. Increasing the number of generated images from 1 to 4 improves AP$_{50}^{novel}$ from 47.4 to 47.9, indicating that additional samples can reduce the randomness of individual generations. However, excessive synthetic images introduce redundant or biased contexts, leading to degraded base-category performance and overall generalization. Meanwhile, GPU memory usage increases substantially from 15GB to 32GB as the number of images grows from 1 to 6. Considering both performance and computational cost, a small number of synthetic images ($num \in [1,4)$) is sufficient, and we adopt $num=1$ as the default setting due to its competitive performance and lowest memory consumption.

% To investigate the optimal number of synthetic images required per novel category to stabilize the visual prototype $p_{vp}$, we evaluate different numbers of generated images on the OV-COCO benchmark, as shown in Fig.~\ref{fig:ab_num}. We report both detection performance and GPU memory usage to analyze the trade-off between prototype stability and computational cost. As the number of synthetic images increases from 1 to 4, AP$_{50}^{novel}$ gradually improves from 47.4 to 47.9, indicating that additional generated samples can help reduce the randomness of a single synthetic image and provide a more stable visual prototype. However, AP$_{50}^{base}$ starts to decrease when the number reaches 4, and both AP$_{50}^{novel}$ and AP$_{50}^{base}$ decline at 6 images. This suggests that excessive synthetic images may introduce redundant or biased generated contexts, which can disturb the balance between base and novel categories rather than further improving generalization.

% Meanwhile, GPU memory usage increases substantially from 15GB to 32GB as the number of synthetic images grows from 1 to 6. Considering both the performance-memory trade-off and the balance between base and novel categories, using a small number of synthetic images per novel category, \textit{i.e.}, $num \in [1,4)$, is sufficient in practice. In our experiments, we adopt $num=1$ as the default setting due to its competitive performance and lowest memory cost.

\begin{table}[t]
\centering
\setlength{\tabcolsep}{14pt}
\captionof{table}{\textbf{Sensitivity analysis of ViT layer groupings.} We evaluate different layer groups used to construct multi-scale FPN features on the OV-COCO benchmark.}
\resizebox{\columnwidth}{!}{
\begin{tabular}{c l>{\columncolor[HTML]{EAEAFF}}ccc}
\toprule
\multicolumn{1}{c}{\multirow{2}{*}{\#}} &
\multicolumn{1}{l}{\multirow{2}{*}{Layer Group}} & \multicolumn{3}{c}{OV-COCO} \\ \cmidrule(l){3-5} 
\multicolumn{1}{c}{} & \multicolumn{1}{c}{} & $\text{AP}_\text{50}^\text{novel}$ & $\text{AP}_\text{50}^\text{base}$ & $\text{AP}_\text{50}^\text{all}$   \\ \midrule
1 & \texttt{[6,10,14,23]}  & \textbf{47.4}      & 63.3               & \underline{59.1} \\
2 & \texttt{[6,10,14,20]}  & 45.8               & 63.1               & 58.6             \\
3 & \texttt{[2,10,14,23]}  & 46.3               & \underline{63.5}   & 58.9             \\
4 & \texttt{[2,6,8,10]}    & 44.2               & 60.7               & 56.4             \\
5 & \texttt{[14,16,18,23]} & 45.2               & \textbf{64.8}      & \textbf{59.8}    \\
6 & \texttt{[5,11,17,23]}  & 46.5               & 63.0               & 58.7             \\
7 & \texttt{[7,11,15,23]}  & \underline{47.0}   & 63.4               & 59.0             \\ \bottomrule
\end{tabular}%
}
\label{tab:layer}
\end{table}

\noindent
\textbf{Sensitivity of ViT layer groupings.}
We further evaluate multiple ViT layer groupings to analyze their influence on multi-scale FPN construction, as shown in Table~\ref{tab:layer}. The evaluated settings cover the following representative choices:
\begin{itemize}
    \item the default balanced grouping following \cite{33_clipself} (\#1);
    \item removal of the final semantic layer (\#2);
    \item an overly shallow early-layer variant (\#3);
    \item shallow- and deep-biased groupings (\#4--\#5);
    \item nearby balanced alternatives (\#6--\#7).
\end{itemize}

% Overall, \#1 achieves the best novel-category performance with 47.4 AP$_{50}^{novel}$, while \#7 obtains comparable results, indicating that PR-OVD is relatively robust to nearby balanced layer choices.
% Specifically, \#2 shows that replacing the final 23rd layer with the 20th layer reduces AP$_{50}^{novel}$, confirming the importance of the final CLIP-level semantic layer for novel-category recognition. \#3 and \#4 suggest that overly shallow features lack sufficient semantic abstraction for open-vocabulary transfer. In contrast, \#5 achieves the highest AP$_{50}^{base}$ but much lower AP$_{50}^{novel}$, indicating that deep-biased features better fit base categories with dense supervision but weaken novel-category generalization. Finally, \#6 and \#7 show that nearby balanced alternatives remain competitive, suggesting that the exact indices are less critical than maintaining balanced coverage of early, middle, and final semantic layers. Since novel-category performance is the primary indicator of open-vocabulary generalization \cite{39_baron}, and \#1 achieves the best AP$_{50}^{novel}$ while maintaining competitive base and overall AP, we adopt \#1, \texttt{[6,10,14,23]}, as the default setting.

Overall, \#1 achieves the best novel-category performance with 47.4 AP$_{50}^{novel}$, while \#7 remains comparable, demonstrating robustness to nearby balanced layer choices. \#2 confirms the importance of the final CLIP semantic layer, whereas \#3 and \#4 show that overly shallow features lack sufficient semantic abstraction. In contrast, \#5 favors base categories but weakens novel-category generalization. The competitive results of \#6 and \#7 further suggest that balanced coverage across early, middle, and final semantic layers is more important than the exact layer indices. Therefore, we adopt \#1, \texttt{[6,10,14,23]}, as the default setting.

\noindent\textbf{Effectiveness of computational efficiency.} To evaluate the computational efficiency of our approach, we measure the training time and GPU memory on OV-COCO with 8*RTX 3090 GPUs. Our PR-OVD takes roughly 7 hours to converge and occupies about 15 GB/GPU of memory, whereas the conventional OVD methods \cite{33_clipself} typically require 5.5 hours and consume 13 GB/GPU, respectively. The slight cost increase stems from the semantic reasoning introduced in our prediction paradigm, which remains acceptable considering the performance gain it brings.

\begin{table}[t]
    \centering
    \setlength{\tabcolsep}{10pt}
    % 为了防止左表标题比右表多一行导致错位，可以统一给标题一个固定高度，或者手动换行
    \caption{\textbf{Transferability results across datasets.} Methods are trained on OV-LVIS and evaluated on COCO and Objects365.}
    \label{table:trans_data}
    \resizebox{\columnwidth}{!}{%
    \begin{tabular}{lcccccc}
    \toprule
    \multirow{2}{*}{Method} & \multicolumn{3}{c}{COCO} & \multicolumn{3}{c}{Objects365} \\
    \cmidrule(lr){2-4} \cmidrule(lr){5-7}
     & AP$_{50}$ & AP$_{75}$ & AP  & AP$_{50}$ & AP$_{75}$ & AP\\
    \midrule
    Supervised  \cite{36_vild}  & 67.6 & 50.9 & 46.5  & 38.6 & 28.0 &  25.6\\
    \midrule
    ViLD \cite{36_vild}    & 55.6 & \underline{39.8} & \underline{36.6}  & 18.2 & 12.6 & 11.8\\
    F-VLM  \cite{38_fvlm}   & 53.1 & 34.6 & 32.5  & 19.2 & 12.6 & 11.9\\
    BARON  \cite{39_baron}  & 55.7 & 39.1 & 36.2  & \underline{21.0} & \textbf{14.5} & \underline{13.6}\\
    LP-OVOD \cite{53_lpovod} & - & - & -  & 18.9 & 13.1 & 12.6\\
    ProxyDet \cite{16_proxydet}   & \underline{57.0} & - & -  & 19.1 & - & -\\
    Det-Agent \cite{79_detagent} & 54.4 & - & 34.7  & 19.2 & - & 12.1
    \\
    \rowcolor[HTML]{EAEAFF}
    \textbf{PR-OVD (Ours)} & \textbf{63.3} & \textbf{42.7} & \textbf{39.9}  & \textbf{24.1} & \underline{14.2} & \textbf{14.0}\\
    \bottomrule
    \end{tabular}%
    }
\end{table}

\subsection{Transferability Analysis}
In this section, we rigorously examine the transferability of our partition-and-refine process. We evaluate its zero-shot generalization across datasets (Tab.\ref{table:trans_data}), assess its potential as a plug-in enhancement for existing OVD frameworks (Tab.\ref{table:trans_method}), and further explore its adaptability to other open-vocabulary tasks by integrating it at the representation level (Tab.\ref{tab:transseg} and \ref{tab:transmultilabel}), demonstrating its robustness and broad applicability.

% \begin{wraptable}{r}{0.5\textwidth}
% \centering
% \vspace{-35pt}
% % \setlength{\tabcolsep}{9pt}
% \captionof{table}{Transferability results across datasets. Methods are trained on OV-LVIS and evaluated on COCO and Objects365 without fine-tuning.}
% \resizebox{0.5\textwidth}{!}{%
% % \setlength{\tabcolsep}{3pt}
% \begin{tabular}{lcccccc}
% \toprule
% \multirow{2}{*}{Method} & \multicolumn{3}{c}{COCO} & \multicolumn{3}{c}{Objects365} \\
% \cmidrule(lr){2-4} \cmidrule(lr){5-7}
%  & AP$_{50}$ & AP$_{75}$ & AP  & AP$_{50}$ & AP$_{75}$ & AP\\
% \midrule
% Supervised  \cite{36_vild}  & 67.6 & 50.9 & 46.5  & 38.6 & 28.0 &  25.6\\
% \midrule
% ViLD \cite{36_vild}    & 55.6 & \underline{39.8} & \underline{36.6}  & 18.2 & 12.6 & 11.8\\
% % DetPro \cite{37_detpro}  & 53.8 & 37.4 & 34.9  & 18.8 & 12.9 & 12.1\\
% F-VLM  \cite{38_fvlm}   & 53.1 & 34.6 & 32.5  & 19.2 & 12.6 & 11.9\\
% BARON  \cite{39_baron}  & 55.7 & 39.1 & 36.2  & \underline{21.0} & \textbf{14.5} & \underline{13.6}\\
% LP-OVOD \cite{53_lpovod} & - & - & -  & 18.9 & 13.1 & 12.6\\
% ProxyDet \cite{16_proxydet}   & \underline{57.0} & - & -  & 19.1 & - & -\\
% \rowcolor[HTML]{EAEAFF}
% \textbf{PR-OVD (Ours)} & \textbf{63.3} & \textbf{42.7} & \textbf{39.9}  & \textbf{24.1} & \underline{14.2} & \textbf{14.0}\\
% \bottomrule
% \end{tabular}%
% }
% \label{table:trans_data}
% \vspace{-16pt}
% \end{wraptable}

% \paragraph{Transfer to other datasets.}
\noindent\textbf{Transfer to other datasets.}
Open-world object detection often encounters substantial domain and vocabulary shifts across datasets. To evaluate the cross-domain and cross-vocabulary generalization of our approach, we directly transfer the detector trained on OV-LVIS to COCO and Objects365 without fine-tuning, replacing only the test-time vocabulary.
% simply replacing the vocabulary with that of the target test set. 
As shown in Tab.\ref{table:trans_data}, our method achieves significant improvements, surpassing the second-best model by \textbf{6.3\%} AP$_{50}$ on COCO. These results highlight the strong generalization capacity of our PR-OVD in diverse open-world scenarios.

\begin{table}[t]
    \centering
    \setlength{\tabcolsep}{16pt}
    % 2. 这里的标题如果比左边短，可以加一个空行 \\ \vspace{...} 撑开，或者手动换行保持一致
    \caption{\textbf{Transferability results of integrating our method into other OVD methods} on OV-COCO.} 
    \label{table:trans_method}
    % 3. 关键：因为右边行数少，通过增加行间距系数，让它在视觉上和左边拉到一样长
    % \renewcommand{\arraystretch}{1} 
    \resizebox{\columnwidth}{!}{%
    \begin{tabular}{l l l l}
    \toprule
    \multirow{2}{*}{Method}               & \multirow{2}{*}{Type}    & \multicolumn{2}{c}{OV-COCO} \\
    \cmidrule(lr){3-4}
    & & $\text{AP}_\text{50}^\text{novel}$ & $\text{AP}_\text{50}^\text{all}$ \\
    \midrule
    F-VLM \cite{38_fvlm}      & CLIP-trans.          & 28.0    & 39.6 \\
    \rowcolor[HTML]{EAEAFF}
    \textbf{+ Ours}                   &       & $\textbf{33.1}^{{\color[HTML]{109e10}\textbf{\small +5.1}}}$ & $\textbf{46.0}^{{\color[HTML]{109e10}\textbf{\small +6.4}}}$ \\
    \midrule
    RALF \cite{15_ralf}       & Pseudo-label.      & 41.3    & 50.9 \\
    \rowcolor[HTML]{EAEAFF}
    \textbf{+ Ours  }                 &       & $\textbf{44.8}^{{\color[HTML]{109e10}\textbf{\small +3.5}}}$ &  $\textbf{55.9}^{{\color[HTML]{109e10}\textbf{\small +5.0}}}$ \\
    \midrule
    CLIPSelf \cite{33_clipself} & CLIP-distill      & 44.3    & 59.0 \\
    \rowcolor[HTML]{EAEAFF}
    \textbf{+ Ours  }                 &        & $\textbf{47.4}^{{\color[HTML]{109e10}\textbf{\small +3.1}}}$ & $\textbf{59.1}^{{\color[HTML]{109e10}\textbf{\small +0.1}}}$ \\
    \bottomrule
    \end{tabular}%
    }
\end{table}

\begin{table}[t]
    \centering
    \renewcommand{\arraystretch}{1.0} % 稍微微调行高以匹配右侧视觉高度
    \setlength{\tabcolsep}{10pt}
    \captionof{table}{\textbf{Performance on Open-Vocabulary Semantic Segmentation.} Trained on COCO-Stuff and evaluated on ADE20K and Pascal-Context.}
    \label{tab:transseg}
    \resizebox{\columnwidth}{!}{%
        \begin{tabular}{lcccc}
            \toprule
            Method & A-150 & A-847 & PC-59 & PC-459 \\ \midrule
            CEL-PL \textsubscript{\textcolor{gray}{TMM'23}} \cite{80_celpl}
            & 29.9 & 9.7 & 55.6 & 12.6 \\
            EBSeg \textsubscript{\textcolor{gray}{CVPR'24}} \cite{55_ebseg} & 30.0 & 11.7 & 56.7 & 17.3 \\
            SED \textsubscript{\textcolor{gray}{CVPR'24}} \cite{56_sed} & 31.6 & 11.4 & 57.3 & 18.6 \\
            CAT-Seg \textsubscript{\textcolor{gray}{CVPR'24}} \cite{57_catseg} & 31.8 & 12.0 & 57.5 & 19.0 \\
            Mask-adapter \textsubscript{\textcolor{gray}{CVPR'25}} \cite{59_maskadapter} & \underline{35.6} & \underline{14.2} & \underline{58.4} & 17.9 \\
            H-CLIP \textsubscript{\textcolor{gray}{CVPR'25}} \cite{58_hclip} & 32.4 & 12.5 & 57.9 & \textbf{19.4} \\
            \rowcolor[HTML]{EAEAFF}
            \textbf{Our PR-OVSS} & \textbf{36.4} & \textbf{14.7} & \textbf{58.7} & \underline{19.1} \\ \bottomrule
        \end{tabular}%
    }
\end{table}

% \paragraph{Transfer to Other OVD Methods.}
\noindent\textbf{Transfer to other OVD methods.}
To evaluate the versatility and scalability of our framework, we integrate it into three representative OVD methods: the CLIP-based method F-VLM~\cite{38_fvlm} and CLIPSelf~\cite{33_clipself}, and the pseudo-label-based method RALF~\cite{15_ralf}.
As illustrated in Tab.\ref{table:trans_method}, incorporating our method consistently boosts performance on the OV-COCO dataset across different baselines. 
% Specifically, we observe AP$_{50}^\text{novel}$ gains of 5.1\% to F-VLM. 
Notably, these methods employ different visual backbones (ResNet-50 and ViT-B/16), yet all benefit from our framework, indicating that the improvements are not attributable to backbone capacity but stem from the proposed partition-and-refine mechanism.

% \paragraph{Transfer to Other Open Vocabulary Tasks.}
\noindent\textbf{Transfer to other open vocabulary tasks.}
Since novel categories are inherent to open-vocabulary tasks, further extend our framework to semantic segmentation and multi-label classification, as shown in Tab.\ref{tab:transseg} and Tab.\ref{tab:transmultilabel}.
By constructing mask–word and image–multi-word alignment spaces, PR-OVD achieves consistent performance gains in both settings, demonstrating its ability to enhance cross-modal understanding across diverse open-vocabulary scenarios.settings, confirming that  enhances cross-modal understanding across diverse open-vocabulary tasks.

\begin{table}[t]
    \centering
    \renewcommand{\arraystretch}{1.0}
    \setlength{\tabcolsep}{6pt}
    \captionof{table}{\textbf{Performance on Open-Vocabulary Multi-Label Classification} on NUS-WIDE. P@3, F1@3 and mAP are reported.}
    \label{tab:transmultilabel}
    \resizebox{\columnwidth}{!}{%
        \begin{tabular}{lccccccc}
            \toprule
            \multirow{2}{*}{Method} & \multicolumn{3}{l}{Zero-shot Learning} & & \multicolumn{3}{l}{G-zero-shot Learning} \\ \cmidrule(lr){2-4} \cmidrule(l){6-8} 
             & P@3 & F1@3 & mAP & & P@3 & F1@3 & mAP \\ \midrule 
            MKT \textsubscript{\textcolor{gray}{AAAI'23}} \cite{60_mkt} & 27.7 & 34.1 & 37.6 & & 35.9 & 22.0 & 18.3 \\
            OVML \textsubscript{\textcolor{gray}{ICME'23}} \cite{61_ovml} & 36.3 & 40.1 & 42.6 & & 32.9 & 18.3 & 14.3 \\
            RAM \textsubscript{\textcolor{gray}{CVPR'25}} \cite{63_ram} & \underline{43.7} & \underline{48.4} & \underline{53.2} & & \underline{38.7} & \underline{23.5} & \underline{20.7} \\
            CoMC \textsubscript{\textcolor{gray}{ICML'24}} \cite{62_comc} & 33.5 & 41.2 & 48.2 & & - & - & - \\
            \rowcolor[HTML]{EAEAFF}
            \textbf{Our PR-OVMLC} & \textbf{44.4} & \textbf{49.3} & \textbf{53.8} & & \textbf{39.2} & \textbf{24.2} & \textbf{21.2} \\ \bottomrule
        \end{tabular}%
    }
\end{table}

\begin{figure}[t]
\centering
% \vspace{-22pt}
\includegraphics[width=0.9\columnwidth]{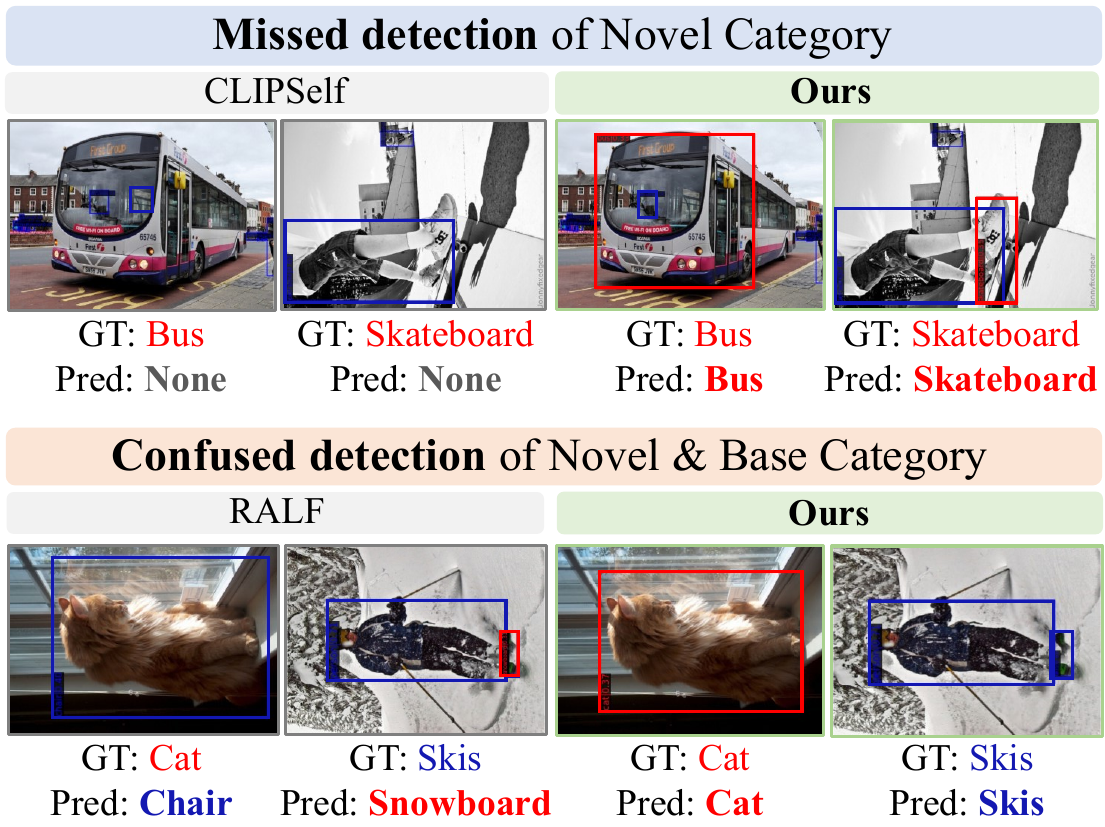}
\caption{\textbf{Visualization of our PR-OVD, CLIPSelf \cite{33_clipself}, and RALF \cite{15_ralf}} on the OV-COCO. Red and blue bounding boxes denote \textcolor[HTML]{FF0000}{novel} and \textcolor[HTML]{1318B1}{base} categories.}
% Red and blue bounding boxes represent novel and base categories, respectively.} 
\label{fig:visual}
% \vspace{-20pt}
\end{figure}

% \begin{figure}[t]
% \centering
% % \vspace{-22pt}
% \includegraphics[width=\columnwidth]{figs/supply_coco.pdf}
% \caption{\textbf{More visualization of our PR-OVD on the OV-COCO.} Red and blue bounding boxes denote \textcolor[HTML]{FF0000}{novel} and \textcolor[HTML]{1318B1}{base} categories.}
% % Red and blue bounding boxes represent novel and base categories, respectively.} 
% \label{fig:visual_coco}
% % \vspace{-20pt}
% \end{figure}

\begin{figure}[t]
\centering
% \vspace{-22pt}
\includegraphics[width=0.9\columnwidth]{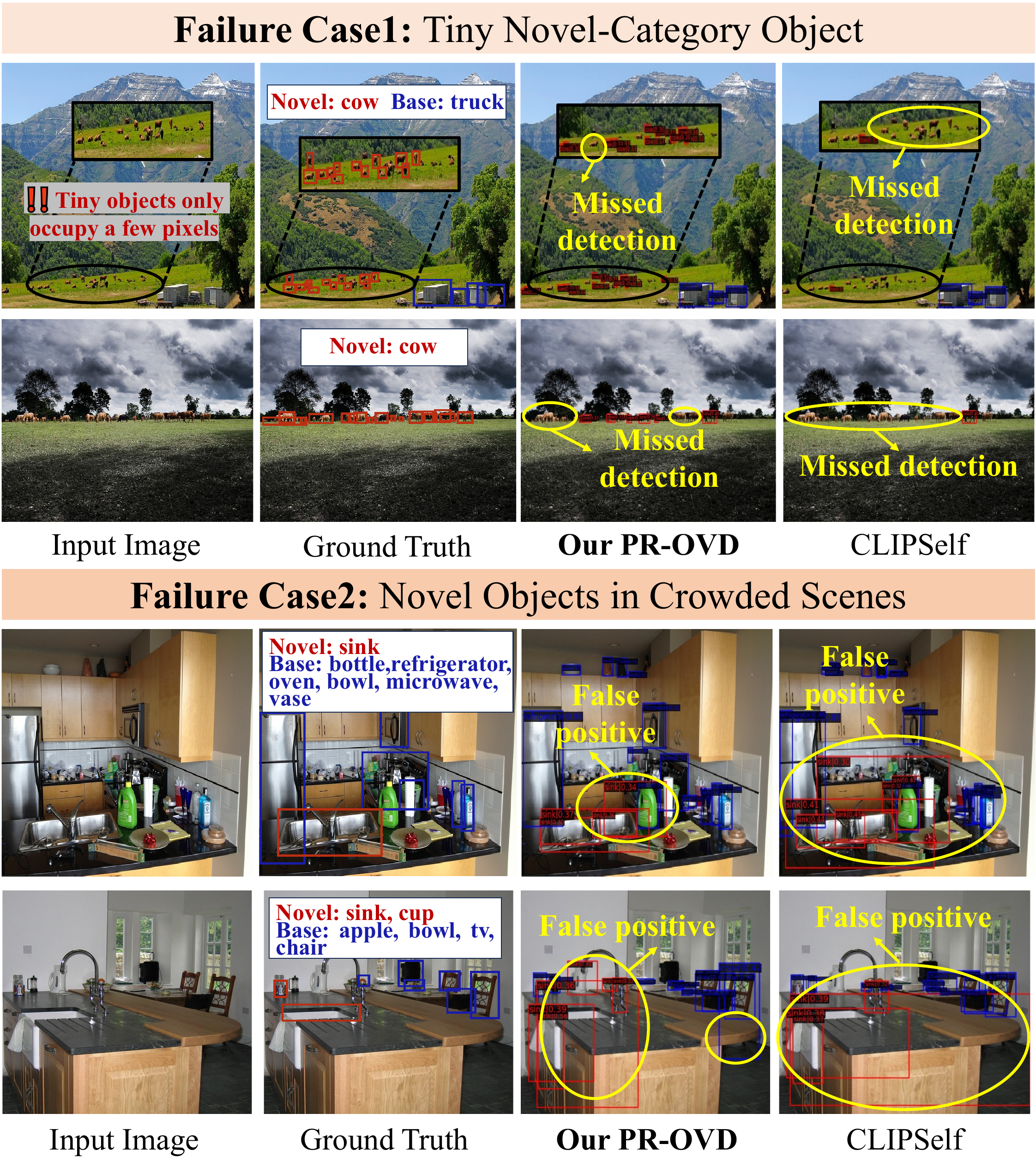}
\caption{\textbf{Representative challenging scenarios for novel-category detection in OVD.} We illustrate two common cases, including tiny novel-category objects and novel objects in crowded scenes, with comparisons between our PR-OVD and existing SoTA method.}
\label{fig:failure}
\end{figure}

\subsection{Visualization and Analysis}
\noindent\textbf{Visualization on OVD task.} To qualitatively evaluate our method, we compare it with two leading baselines on OV-COCO: CLIPSelf \cite{33_clipself} and RALF~\cite{15_ralf}.
As shown in Fig.\ref{fig:visual},
% red and blue boxes denote novel and base categories, respectively. 
prior methods commonly suffer from: (1) missed detections of novel category due to occlusion, viewpoint variation (\textit{e.g.}, bus), or weak visual signals (\textit{e.g.}, skateboard), and (2) confused detections between novel and base classes due to high visual similarity(\textit{e.g.}, skis vs. snowboard).
In contrast, our method alleviates these issues by explicitly reserving dedicated regions for novel categories and refining category-level geometry within the partitioned space. This leads to more robust detection, especially in visually ambiguous scenarios. 
% More visualizations are provided in Fig.\ref{fig:visual_coco}.

\noindent
\textbf{Failure case analysis.} We analyze two representative challenging scenarios commonly encountered in OVD, as shown in Fig.~\ref{fig:failure}: tiny novel-category objects and novel objects in crowded scenes. Tiny objects provide limited visual evidence and are easily overwhelmed by background patterns, causing missed detections. Crowded scenes introduce severe instance interference and ambiguous object boundaries, leading to false positives or localization errors. These failures reflect inherent challenges of OVD, where novel categories lack direct region-level supervision. Despite these challenges, PR-OVD still achieves more robust predictions than CLIPSelf~\cite{33_clipself}. Future work will explore stronger multi-scale representations, object-centric generation, and instance-level reasoning to further improve challenging open-vocabulary scenarios.

% \begin{figure}[t]
% \centering
% % \vspace{-22pt}
% \includegraphics[width=\columnwidth]{figs/visual_ade.pdf}
% \caption{\textbf{Visualization of our PR-OVSS on ADE-20K (A-150),} including the original image, ground-truth, and the zero-shot segmentation results.}
% % Red and blue bounding boxes represent novel and base categories, respectively.} 
% \label{fig:visual_ade}
% % \vspace{-20pt}
% \end{figure}

% \begin{figure}[t]
% \centering
% % \vspace{-22pt}
% \includegraphics[width=\columnwidth]{figs/visual_pc.pdf}
% \caption{\textbf{Visualization of our PR-OVSS on Pascal-Context (PC-59).}}
% % Red and blue bounding boxes represent novel and base categories, respectively.} 
% \label{fig:visual_pc}
% % \vspace{-20pt}
% \end{figure}

% \noindent\textbf{Visualization on OVSS task.} We further extend our method to open-vocabulary semantic segmentation by constructing mask-level language–vision alignment and following the MaskAdapter training protocol \cite{59_maskadapter}. The model is trained on COCO-Stuff and directly evaluated on ADE20K and Pascal-Context without any dataset-specific tuning.
% As shown in Fig.\ref{fig:visual_ade} and Fig.\ref{fig:visual_pc}, PR-OVSS demonstrates excellent zero-shot transfer capabilities, generating semantically consistent masks and maintaining clear object boundaries across different scenes. Even for unseen categories, it exhibits stable region consistency, validating the effectiveness of our method.

\section{Conclusions}
\label{sec:conclusion}

In this work, we propose a Partition-and-Refine framework for open-vocabulary object detection (PR-OVD), explicitly reserving dedicated capacity for novel categories to prevent semantic collapse and calibration imbalance.
As its core, PR-OVD leverages a coarse partition and category-level refinement module to prevent novel representations from being absorbed by dominant base clusters and sharpen inter-class boundary. Additionally, we design a cognitively inspired structure prompt that generates visual cues to enhance novel category representation.
% Our method achieves state-of-the-art performance on all OVD benchmarks and integrates seamlessly into existing OVD pipelines.
Beyond detection, PR-OVD naturally transfers to other open-vocabulary tasks, achieving performance improvements while offering a fresh perspective for open vocabulary understanding.

% \section*{Acknowledgments}
% This should be a simple paragraph before the References to thank those individuals and institutions who have supported your work on this article.

% {\appendix[Proof of the Zonklar Equations]
% Use $\backslash${\tt{appendix}} if you have a single appendix:
% Do not use $\backslash${\tt{section}} anymore after $\backslash${\tt{appendix}}, only $\backslash${\tt{section*}}.
% If you have multiple appendixes use $\backslash${\tt{appendices}} then use $\backslash${\tt{section}} to start each appendix.
% You must declare a $\backslash${\tt{section}} before using any $\backslash${\tt{subsection}} or using $\backslash${\tt{label}} ($\backslash${\tt{appendices}} by itself
%  starts a section numbered zero.)}

\bibliographystyle{IEEEtran}
\bibliography{tmm2026}

@String(ICML  = {Int. Conf. Mach. Learn.})

@String(ICLR  = {Int. Conf. Learn. Represent.})

@String(AAAI  = {AAAI})

@String(ICME  = {Int. Conf. Multimedia and Expo})

@String(ICML  = {ICML})

@String(ICLR  = {ICLR})

@String(ICME  =	{ICME})

@article{2,
  title={Reasoning multi-agent behavioral topology for interactive autonomous driving},
  author={Liu, Haochen and Chen, Li and Qiao, Yu and Lv, Chen and Li, Hongyang},
  journal={Advances in Neural Information Processing Systems},
  volume={37},
  pages={92605--92637},
  year={2024}
}

@inproceedings{6_mrcnn,
  title={Mask r-cnn},
  author={He, Kaiming and Gkioxari, Georgia and Doll{\'a}r, Piotr and Girshick, Ross},
  booktitle={Proceedings of the IEEE international conference on computer vision},
  pages={2961--2969},
  year={2017}
}

@inproceedings{8_yolo,
  title={You only look once: Unified, real-time object detection},
  author={Redmon, Joseph and Divvala, Santosh and Girshick, Ross and Farhadi, Ali},
  booktitle={Proceedings of the IEEE conference on computer vision and pattern recognition},
  pages={779--788},
  year={2016}
}

@inproceedings{10_ovrcnn,
  title={Open-vocabulary object detection using captions},
  author={Zareian, Alireza and Rosa, Kevin Dela and Hu, Derek Hao and Chang, Shih-Fu},
  booktitle={Proceedings of the IEEE/CVF conference on computer vision and pattern recognition},
  pages={14393--14402},
  year={2021}
}

@inproceedings{11_regionclip,
  title={Regionclip: Region-based language-image pretraining},
  author={Zhong, Yiwu and Yang, Jianwei and Zhang, Pengchuan and Li, Chunyuan and Codella, Noel and Li, Liunian Harold and Zhou, Luowei and Dai, Xiyang and Yuan, Lu and Li, Yin and others},
  booktitle={Proceedings of the IEEE/CVF conference on computer vision and pattern recognition},
  pages={16793--16803},
  year={2022}
}

@inproceedings{14_sasdet,
  title={Taming self-training for open-vocabulary object detection},
  author={Zhao, Shiyu and Schulter, Samuel and Zhao, Long and Zhang, Zhixing and Suh, Yumin and Chandraker, Manmohan and Metaxas, Dimitris N and others},
  booktitle={Proceedings of the IEEE/CVF Conference on Computer Vision and Pattern Recognition},
  pages={13938--13947},
  year={2024}
}

@inproceedings{15_ralf,
  title={Retrieval-augmented open-vocabulary object detection},
  author={Kim, Jooyeon and Cho, Eulrang and Kim, Sehyung and Kim, Hyunwoo J},
  booktitle={Proceedings of the IEEE/CVF Conference on Computer Vision and Pattern Recognition},
  pages={17427--17436},
  year={2024}
}

@inproceedings{16_proxydet,
  title={Proxydet: Synthesizing proxy novel classes via classwise mixup for open-vocabulary object detection},
  author={Jeong, Joonhyun and Park, Geondo and Yoo, Jayeon and Jung, Hyungsik and Kim, Heesu},
  booktitle={Proceedings of the AAAI Conference on Artificial Intelligence},
  volume={38},
  number={3},
  pages={2462--2470},
  year={2024}
}

@article{17_survey,
  title={A survey on open-vocabulary detection and segmentation: Past, present, and future},
  author={Zhu, Chaoyang and Chen, Long},
  journal={IEEE Transactions on Pattern Analysis and Machine Intelligence},
  volume={46},
  number={12},
  pages={8954--8975},
  year={2024},
  publisher={IEEE}
}

@article{18_detclip,
  title={Detclip: Dictionary-enriched visual-concept paralleled pre-training for open-world detection},
  author={Yao, Lewei and Han, Jianhua and Wen, Youpeng and Liang, Xiaodan and Xu, Dan and Zhang, Wei and Li, Zhenguo and Xu, Chunjing and Xu, Hang},
  journal={Advances in Neural Information Processing Systems},
  volume={35},
  pages={9125--9138},
  year={2022}
}

@inproceedings{23_cora,
  title={Cora: Adapting clip for open-vocabulary detection with region prompting and anchor pre-matching},
  author={Wu, Xiaoshi and Zhu, Feng and Zhao, Rui and Li, Hongsheng},
  booktitle={Proceedings of the IEEE/CVF conference on computer vision and pattern recognition},
  pages={7031--7040},
  year={2023}
}

@article{25_retrieval,
  title={Retrieval-augmented generation for knowledge-intensive nlp tasks},
  author={Lewis, Patrick and Perez, Ethan and Piktus, Aleksandra and Petroni, Fabio and Karpukhin, Vladimir and Goyal, Naman and K{\"u}ttler, Heinrich and Lewis, Mike and Yih, Wen-tau and Rockt{\"a}schel, Tim and others},
  journal={Advances in neural information processing systems},
  volume={33},
  pages={9459--9474},
  year={2020}
}

@inproceedings{27_clip,
  title={Learning transferable visual models from natural language supervision},
  author={Radford, Alec and Kim, Jong Wook and Hallacy, Chris and Ramesh, Aditya and Goh, Gabriel and Agarwal, Sandhini and Sastry, Girish and Askell, Amanda and Mishkin, Pamela and Clark, Jack and others},
  booktitle={International conference on machine learning},
  pages={8748--8763},
  year={2021},
  organization={PmLR}
}

@inproceedings{28_moco,
  title={Momentum contrast for unsupervised visual representation learning},
  author={He, Kaiming and Fan, Haoqi and Wu, Yuxin and Xie, Saining and Girshick, Ross},
  booktitle={Proceedings of the IEEE/CVF conference on computer vision and pattern recognition},
  pages={9729--9738},
  year={2020}
}

@inproceedings{29_simclr,
  title={A simple framework for contrastive learning of visual representations},
  author={Chen, Ting and Kornblith, Simon and Norouzi, Mohammad and Hinton, Geoffrey},
  booktitle={International conference on machine learning},
  pages={1597--1607},
  year={2020},
  organization={PmLR}
}

@inproceedings{30_detco,
  title={Detco: Unsupervised contrastive learning for object detection},
  author={Xie, Enze and Ding, Jian and Wang, Wenhai and Zhan, Xiaohang and Xu, Hang and Sun, Peize and Li, Zhenguo and Luo, Ping},
  booktitle={Proceedings of the IEEE/CVF international conference on computer vision},
  pages={8392--8401},
  year={2021}
}

@article{33_clipself,
  title={Clipself: Vision transformer distills itself for open-vocabulary dense prediction},
  author={Wu, Size and Zhang, Wenwei and Xu, Lumin and Jin, Sheng and Li, Xiangtai and Liu, Wentao and Loy, Chen Change},
  journal={arXiv preprint arXiv:2310.01403},
  year={2023}
}

@article{34_infonce,
  title={Representation learning with contrastive predictive coding},
  author={Oord, Aaron van den and Li, Yazhe and Vinyals, Oriol},
  journal={arXiv preprint arXiv:1807.03748},
  year={2018}
}

@inproceedings{35_procontrast,
  author       = {Junnan Li and
                  Pan Zhou and
                  Caiming Xiong and
                  Steven C. H. Hoi},
  title        = {Prototypical Contrastive Learning of Unsupervised Representations},
  booktitle    = {9th International Conference on Learning Representations, {ICLR} 2021,
                  Virtual Event, Austria, May 3-7, 2021},
  year         = {2021},
  bibsource    = {dblp computer science bibliography, https://dblp.org}
}

@inproceedings{36_vild,
  author       = {Xiuye Gu and
                  Tsung{-}Yi Lin and
                  Weicheng Kuo and
                  Yin Cui},
  title        = {Open-vocabulary Object Detection via Vision and Language Knowledge
                  Distillation},
  booktitle    = {The Tenth International Conference on Learning Representations, {ICLR}
                  2022, Virtual Event, April 25-29, 2022},
  year         = {2022},
  bibsource    = {dblp computer science bibliography, https://dblp.org}
}

@inproceedings{38_fvlm,
  author       = {Weicheng Kuo and
                  Yin Cui and
                  Xiuye Gu and
                  A. J. Piergiovanni and
                  Anelia Angelova},
  title        = {F-vlm: Open-Vocabulary Object Detection upon Frozen Vision and Language Models},
  booktitle    = {The Eleventh International Conference on Learning Representations,
                  {ICLR} 2023, Kigali, Rwanda, May 1-5, 2023},
  year         = {2023},
  bibsource    = {dblp computer science bibliography, https://dblp.org}
}

@inproceedings{39_baron,
  title={Aligning bag of regions for open-vocabulary object detection},
  author={Wu, Size and Zhang, Wenwei and Jin, Sheng and Liu, Wentao and Loy, Chen Change},
  booktitle={Proceedings of the IEEE/CVF conference on computer vision and pattern recognition},
  pages={15254--15264},
  year={2023}
}

@inproceedings{40_coco,
  title={Microsoft coco: Common objects in context},
  author={Lin, Tsung-Yi and Maire, Michael and Belongie, Serge and Hays, James and Perona, Pietro and Ramanan, Deva and Doll{\'a}r, Piotr and Zitnick, C Lawrence},
  booktitle={European conference on computer vision},
  pages={740--755},
  year={2014},
  organization={Springer}
}

@inproceedings{41_lvis,
  title={Lvis: A dataset for large vocabulary instance segmentation},
  author={Gupta, Agrim and Dollar, Piotr and Girshick, Ross},
  booktitle={Proceedings of the IEEE/CVF conference on computer vision and pattern recognition},
  pages={5356--5364},
  year={2019}
}

@inproceedings{42_fpn,
  title={Feature pyramid networks for object detection},
  author={Lin, Tsung-Yi and Doll{\'a}r, Piotr and Girshick, Ross and He, Kaiming and Hariharan, Bharath and Belongie, Serge},
  booktitle={Proceedings of the IEEE conference on computer vision and pattern recognition},
  pages={2117--2125},
  year={2017}
}

@inproceedings{43_rovit,
  title={Region-aware pretraining for open-vocabulary object detection with vision transformers},
  author={Kim, Dahun and Angelova, Anelia and Kuo, Weicheng},
  booktitle={Proceedings of the IEEE/CVF conference on computer vision and pattern recognition},
  pages={11144--11154},
  year={2023}
}

@inproceedings{44_cfmvit,
  title={Contrastive feature masking open-vocabulary vision transformer},
  author={Kim, Dahun and Angelova, Anelia and Kuo, Weicheng},
  booktitle={Proceedings of the IEEE/CVF International Conference on Computer Vision},
  pages={15602--15612},
  year={2023}
}

@inproceedings{46_triple,
  title={Vision-language pre-training with triple contrastive learning},
  author={Yang, Jinyu and Duan, Jiali and Tran, Son and Xu, Yi and Chanda, Sampath and Chen, Liqun and Zeng, Belinda and Chilimbi, Trishul and Huang, Junzhou},
  booktitle={Proceedings of the IEEE/CVF conference on computer vision and pattern recognition},
  pages={15671--15680},
  year={2022}
}

@inproceedings{47_siccad,
  title={Simple image-level classification improves open-vocabulary object detection},
  author={Fang, Ruohuan and Pang, Guansong and Bai, Xiao},
  booktitle={Proceedings of the AAAI conference on artificial intelligence},
  volume={38},
  number={2},
  pages={1716--1725},
  year={2024}
}

@inproceedings{48_lbp,
  title={Learning background prompts to discover implicit knowledge for open vocabulary object detection},
  author={Li, Jiaming and Zhang, Jiacheng and Li, Jichang and Li, Ge and Liu, Si and Lin, Liang and Li, Guanbin},
  booktitle={Proceedings of the IEEE/CVF Conference on Computer Vision and Pattern Recognition},
  pages={16678--16687},
  year={2024}
}

@inproceedings{49_vldet,
  author       = {Chuang Lin and
                  Peize Sun and
                  Yi Jiang and
                  Ping Luo and
                  Lizhen Qu and
                  Gholamreza Haffari and
                  Zehuan Yuan and
                  Jianfei Cai},
  title        = {Learning Object-Language Alignments for Open-Vocabulary Object Detection},
  booktitle    = {The Eleventh International Conference on Learning Representations,
                  {ICLR} 2023, Kigali, Rwanda, May 1-5, 2023},
  year         = {2023},
  bibsource    = {dblp computer science bibliography, https://dblp.org}
}

@article{50_cognition,
  title={Knowledge across reference frames: Cognitive maps and image spaces},
  author={Bottini, Roberto and Doeller, Christian F},
  journal={Trends in Cognitive Sciences},
  volume={24},
  number={8},
  pages={606--619},
  year={2020},
  publisher={Elsevier}
}

@inproceedings{51_cake,
  title={Cake: Category aware knowledge extraction for open-vocabulary object detection},
  author={Ma, Shiyuan and Qian, Donglin and Ye, Kai and Zhang, Shengchuan},
  booktitle={Proceedings of the AAAI Conference on Artificial Intelligence},
  volume={39},
  number={6},
  pages={5982--5990},
  year={2025}
}

@inproceedings{52_cckt,
  author       = {Chuhan Zhang and
                  Chaoyang Zhu and
                  Pingcheng Dong and
                  Long Chen and
                  Dong Zhang},
  title        = {Cyclic Contrastive Knowledge Transfer for Open-Vocabulary Object Detection},
  booktitle    = {The Thirteenth International Conference on Learning Representations,
                  {ICLR} 2025, Singapore, April 24-28, 2025},
  publisher    = {OpenReview.net},
  year         = {2025},
  bibsource    = {dblp computer science bibliography, https://dblp.org}
}

@inproceedings{53_lpovod,
  title={LP-OVOD: Open-vocabulary object detection by linear probing},
  author={Pham, Chau and Vu, Truong and Nguyen, Khoi},
  booktitle={Proceedings of the IEEE/CVF Winter Conference on Applications of Computer Vision},
  pages={779--788},
  year={2024}
}

@inproceedings{55_ebseg,
  title={Open-vocabulary semantic segmentation with image embedding balancing},
  author={Shan, Xiangheng and Wu, Dongyue and Zhu, Guilin and Shao, Yuanjie and Sang, Nong and Gao, Changxin},
  booktitle={Proceedings of the IEEE/CVF Conference on Computer Vision and Pattern Recognition},
  pages={28412--28421},
  year={2024}
}

@inproceedings{56_sed,
  title={Sed: A simple encoder-decoder for open-vocabulary semantic segmentation},
  author={Xie, Bin and Cao, Jiale and Xie, Jin and Khan, Fahad Shahbaz and Pang, Yanwei},
  booktitle={Proceedings of the IEEE/CVF conference on computer vision and pattern recognition},
  pages={3426--3436},
  year={2024}
}

@inproceedings{57_catseg,
  title={Cat-seg: Cost aggregation for open-vocabulary semantic segmentation},
  author={Cho, Seokju and Shin, Heeseong and Hong, Sunghwan and Arnab, Anurag and Seo, Paul Hongsuck and Kim, Seungryong},
  booktitle={Proceedings of the IEEE/CVF Conference on Computer Vision and Pattern Recognition},
  pages={4113--4123},
  year={2024}
}

@inproceedings{58_hclip,
  title={Parameter-efficient Fine-tuning in Hyperspherical Space for Open-vocabulary Semantic Segmentation},
  author={Peng, Zelin and Xu, Zhengqin and Zeng, Zhilin and Huang, Yu and Wang, Yaoming and Shen, Wei},
  booktitle={Proceedings of the Computer Vision and Pattern Recognition Conference},
  pages={15009--15020},
  year={2025}
}

@inproceedings{59_maskadapter,
  title={Mask-Adapter: The Devil is in the Masks for Open-Vocabulary Segmentation},
  author={Li, Yongkang and Cheng, Tianheng and Feng, Bin and Liu, Wenyu and Wang, Xinggang},
  booktitle={Proceedings of the Computer Vision and Pattern Recognition Conference},
  pages={14998--15008},
  year={2025}
}

@inproceedings{60_mkt,
  title={Open-vocabulary multi-label classification via multi-modal knowledge transfer},
  author={He, Sunan and Guo, Taian and Dai, Tao and Qiao, Ruizhi and Shu, Xiujun and Ren, Bo and Xia, Shu-Tao},
  booktitle={Proceedings of the AAAI conference on artificial intelligence},
  volume={37},
  number={1},
  pages={808--816},
  year={2023}
}

@inproceedings{61_ovml,
  title={Open-vocabulary multi-label image classification with pretrained vision-language model},
  author={Dao, Son D and Huynh, Dat and Zhao, He and Phung, Dinh and Cai, Jianfei},
  booktitle={2023 IEEE International Conference on Multimedia and Expo (ICME)},
  pages={2135--2140},
  year={2023},
  organization={IEEE}
}

@inproceedings{62_comc,
  title={Language-driven cross-modal classifier for zero-shot multi-label image recognition},
  author={Liu, Yicheng and Wen, Jie and Liu, Chengliang and Fang, Xiaozhao and Li, Zuoyong and Xu, Yong and Zhang, Zheng},
  booktitle={Forty-first International Conference on Machine Learning},
  year={2024}
}

@inproceedings{63_ram,
  title={Recover and Match: Open-Vocabulary Multi-Label Recognition through Knowledge-Constrained Optimal Transport},
  author={Tan, Hao and Tan, Zichang and Li, Jun and Liu, Ajian and Wan, Jun and Lei, Zhen},
  booktitle={Proceedings of the Computer Vision and Pattern Recognition Conference},
  pages={4650--4660},
  year={2025}
}

@inproceedings{64_cocostuff,
  title={Coco-stuff: Thing and stuff classes in context},
  author={Caesar, Holger and Uijlings, Jasper and Ferrari, Vittorio},
  booktitle={Proceedings of the IEEE conference on computer vision and pattern recognition},
  pages={1209--1218},
  year={2018}
}

@inproceedings{65_nuswide,
  title={Nus-wide: a real-world web image database from national university of singapore},
  author={Chua, Tat-Seng and Tang, Jinhui and Hong, Richang and Li, Haojie and Luo, Zhiping and Zheng, Yantao},
  booktitle={Proceedings of the ACM international conference on image and video retrieval},
  pages={1--9},
  year={2009}
}

@inproceedings{66_pascalcontext,
  title={The role of context for object detection and semantic segmentation in the wild},
  author={Mottaghi, Roozbeh and Chen, Xianjie and Liu, Xiaobai and Cho, Nam-Gyu and Lee, Seong-Whan and Fidler, Sanja and Urtasun, Raquel and Yuille, Alan},
  booktitle={Proceedings of the IEEE conference on computer vision and pattern recognition},
  pages={891--898},
  year={2014}
}

@article{67_ade,
  title={Semantic understanding of scenes through the ade20k dataset},
  author={Zhou, Bolei and Zhao, Hang and Puig, Xavier and Xiao, Tete and Fidler, Sanja and Barriuso, Adela and Torralba, Antonio},
  journal={International Journal of Computer Vision},
  volume={127},
  number={3},
  pages={302--321},
  year={2019},
  publisher={Springer}
}

@article{72_human1,
  author       = {Jitendra Parmar and
                  Satyendra Singh Chouhan and
                  Vaskar Raychoudhury},
  title        = {A machine learning based framework to identify unseen classes in open-world
                  text classification},
  journal      = {Inf. Process. Manag.},
  volume       = {60},
  number       = {2},
  pages        = {103214},
  year         = {2023},
  bibsource    = {dblp computer science bibliography, https://dblp.org}
}

@article{73_human2,
  title={An unexpected sequence of events: mismatch detection in the human hippocampus},
  author={Kumaran, Dharshan and Maguire, Eleanor A},
  journal={PLoS biology},
  volume={4},
  number={12},
  pages={e424},
  year={2006},
  publisher={Public Library of Science San Francisco, USA}
}

@inproceedings{74_cc3m,
  title={Conceptual captions: A cleaned, hypernymed, image alt-text dataset for automatic image captioning},
  author={Sharma, Piyush and Ding, Nan and Goodman, Sebastian and Soricut, Radu},
  booktitle={Proceedings of the 56th Annual Meeting of the Association for Computational Linguistics (Volume 1: Long Papers)},
  pages={2556--2565},
  year={2018}
}

@article{75_caption,
  title={Microsoft coco captions: Data collection and evaluation server},
  author={Chen, Xinlei and Fang, Hao and Lin, Tsung-Yi and Vedantam, Ramakrishna and Gupta, Saurabh and Doll{\'a}r, Piotr and Zitnick, C Lawrence},
  journal={arXiv preprint arXiv:1504.00325},
  year={2015}
}

@article{76_gpt4o,
  title={Gpt-4 technical report},
  author={Achiam, Josh and Adler, Steven and Agarwal, Sandhini and Ahmad, Lama and Akkaya, Ilge and Aleman, Florencia Leoni and Almeida, Diogo and Altenschmidt, Janko and Altman, Sam and Anadkat, Shyamal and others},
  journal={arXiv preprint arXiv:2303.08774},
  year={2023}
}

@inproceedings{77_sd3.5,
  author       = {Patrick Esser and
                  Sumith Kulal and
                  Andreas Blattmann and
                  Rahim Entezari and
                  Jonas M{\"{u}}ller and
                  Harry Saini and
                  Yam Levi and
                  Dominik Lorenz and
                  Axel Sauer and
                  Frederic Boesel and
                  Dustin Podell and
                  Tim Dockhorn and
                  Zion English and
                  Robin Rombach},
  editor       = {Ruslan Salakhutdinov and
                  Zico Kolter and
                  Katherine A. Heller and
                  Adrian Weller and
                  Nuria Oliver and
                  Jonathan Scarlett and
                  Felix Berkenkamp},
  title        = {Scaling Rectified Flow Transformers for High-Resolution Image Synthesis},
  booktitle    = {Forty-first International Conference on Machine Learning, {ICML} 2024,
                  Vienna, Austria, July 21-27, 2024},
  series       = {Proceedings of Machine Learning Research},
  volume       = {235},
  pages        = {12606--12633},
  publisher    = {{PMLR} / OpenReview.net},
  year         = {2024},
  bibsource    = {dblp computer science bibliography, https://dblp.org}
}

@article{79_detagent,
  title={Det-Agent: Open-Vocabulary Object Localization and Detection with Reinforcement Learning Agent},
  author={Zhang, Ruisong and Wu, Xin-Jian and Wang, Chuang and Liu, Cheng-Lin},
  journal={IEEE Transactions on Multimedia},
  year={2026},
  publisher={IEEE}
}

@article{80_celpl,
  title={Class enhancement losses with pseudo labels for open-vocabulary semantic segmentation},
  author={Dao, Son Duy and Shi, Hengcan and Phung, Dinh and Cai, Jianfei},
  journal={IEEE Transactions on Multimedia},
  volume={26},
  pages={8442--8453},
  year={2023},
  publisher={IEEE}
}

@article{81_uovn,
  title={Unified open-vocabulary dense visual prediction},
  author={Shi, Hengcan and Hayat, Munawar and Cai, Jianfei},
  journal={IEEE Transactions on Multimedia},
  volume={26},
  pages={8704--8716},
  year={2024},
  publisher={IEEE}
}

@article{82_tmmseg,
  title={OV-BIS: Open-Vocabulary Boundary Guide Zero-Shot 3D Instance Segmentation},
  author={Yi, Tinghao and Wang, Shaohu and Zhang, Zhengtao and Wang, Changwei and Yan, Dongming and Xu, Rongtao and Chen, Enhong},
  journal={IEEE Transactions on Multimedia},
  year={2025},
  publisher={IEEE}
}

@article{83_tmmvideo,
  title={Open-vocabulary multi-object tracking with domain generalized and temporally adaptive features},
  author={Li, Run and Zhang, Dawei and Wang, Yanchao and Jiang, Yunliang and Zheng, Zhonglong and Jeon, Sang-Woon and Wang, Hua},
  journal={IEEE Transactions on Multimedia},
  year={2025},
  publisher={IEEE}
}

@article{84_tmmvideo2,
  title={Towards open-vocabulary video semantic segmentation},
  author={Li, Xinhao and Liu, Yun and Sun, Guolei and Wu, Min and Zhang, Le and Zhu, Ce},
  journal={IEEE Transactions on Multimedia},
  year={2025},
  publisher={IEEE}
}

@article{85_tmmautodive,
  title={Cas-OVD: Cascaded Open-Vocabulary Detection of Small Objects Using Multi-Refined Region Proposal Network in Autonomous Driving},
  author={Fang, Zhenyu and Wu, Yulong and Ren, Jinchang and Zheng, Jiangbin and Yan, Yijun and Zhang, Lixiang},
  journal={IEEE Transactions on Multimedia},
  volume={28},
  pages={757--771},
  year={2025},
  publisher={IEEE}
}

@article{86_qwen,
  title={Qwen-image technical report},
  author={Wu, Chenfei and Li, Jiahao and Zhou, Jingren and Lin, Junyang and Gao, Kaiyuan and Yan, Kun and Yin, Sheng-ming and Bai, Shuai and Xu, Xiao and Chen, Yilei and others},
  journal={arXiv preprint arXiv:2508.02324},
  year={2025}
}

@article{87_llama,
  title={The llama 3 herd of models},
  author={Grattafiori, Aaron and Dubey, Abhimanyu and Jauhri, Abhinav and Pandey, Abhinav and Kadian, Abhishek and Al-Dahle, Ahmad and Letman, Aiesha and Mathur, Akhil and Schelten, Alan and Vaughan, Alex and others},
  journal={arXiv preprint arXiv:2407.21783},
  year={2024}
}

@inproceedings{chi1,
  title={Plug-in feedback self-adaptive attention in clip for training-free open-vocabulary segmentation},
  author={Chi, Zhixiang and Wu, Yanan and Gu, Li and Liu, Huan and Wang, Ziqiang and Zhang, Yang and Wang, Yang and Plataniotis, Konstantinos},
  booktitle={Proceedings of the IEEE/CVF international conference on computer vision},
  pages={22815--22825},
  year={2025}
}

@inproceedings{chi2,
  title={Bridging the Modality Gap in Compositional Zero-Shot Learning via Sparse Alignment and Unimodal Memory Bank},
  author={Zhang, Yang and Chi, Zhixiang and Yan, Xudong and Wang, Yang and Feng, Songhe},
  booktitle={Proceedings of the IEEE/CVF Conference on Computer Vision and Pattern Recognition},
  pages={5553--5563},
  year={2026}
}

@article{chi3,
  title={TALON: Test-time Adaptive Learning for On-the-Fly Category Discovery},
  author={Wu, Yanan and Yan, Yuhan and Chen, Tailai and Chi, Zhixiang and Wu, ZiZhang and Jin, Yi and Wang, Yang and Li, Zhenbo},
  journal={arXiv preprint arXiv:2603.08075},
  year={2026}
}

\vfill

\end{document}